\documentclass[9pt,twocolumn,twoside]{osajnl}
\journal{ol} 
\setboolean{shortarticle}{false}

\title{System Design of the Ultra Mobility Vehicle: \\A Driving, Balancing, and Jumping Bicycle Robot}

\author[1,*]{Benjamin Bokser}
\author[1]{Daniel J. Gonzalez}
\author[1]{Aaron Preston}
\author[1]{Alex Bahner}
\author[2]{Annika Wollschläger}
\author[1]{Arianna Ilvonen}
\author[1]{Asa Eckert-Erdheim}
\author[1]{Ashwin Khadke}
\author[1]{Bilal Hammoud}
\author[1]{Dean Molinaro}
\author[2]{Fabian Jenelten}
\author[1]{Henry Mayne}
\author[3]{Howie Choset}
\author[2]{Igor Bogoslavskyi}
\author[1]{Itic Tinman}
\author[1]{James Tigue}
\author[2]{Jan Preisig}
\author[1]{Kaiyu Zheng}
\author[1]{Kenny Sharma}
\author[1]{Kim Ang}
\author[1]{Laura Lee}
\author[1]{Liana Margolese}
\author[1]{Nicole Lin}
\author[1]{Oscar Frias}
\author[1]{Paul Drews}
\author[1]{Ravi Boggavarapu}
\author[1]{Rick Burnham}
\author[1]{Samuel Zapolsky}
\author[4]{Sangbae Kim}
\author[1]{Scott Biddlestone}
\author[1]{Sean Mayorga}
\author[1]{Shamel Fahmi}
\author[1,*]{Surya P. N. Singh}
\author[1]{Tyler McCollum}
\author[1]{Velin Dimitrov}
\author[1]{William Moyne}
\author[1]{Yu-Ming Chen}
\author[1,\dag]{David Perry}
\author[1,\dag]{Farbod Farshidian}
\author[2,\dag]{Marco Hutter}
\author[1,\dag]{Al Rizzi}
\author[1,\dag]{Gabe Nelson}

\affil[1]{Robotics and AI Institute (RAI), Cambridge MA, USA}
\affil[2]{Robotics and AI Institute (RAI), Zurich, Switzerland}
\affil[3]{Robotics Institute, Carnegie Mellon University, Pittsburgh PA, USA}
\affil[4]{Mechanical Engineering Department, Massachusetts Institute of Technology, Cambridge MA, USA}
\affil[$\dagger$]{Project Lead}
\affil[*]{Corresponding authors: \href{mailto:bbokser@rai-inst.com}{bbokser@rai-inst.com}, \href{mailto:ssingh@rai-inst.com}{ssingh@rai-inst.com} }

\usepackage[]{units} % for units
\usepackage{xspace} % to handle spaces after new commands
\usepackage{hyperref} % Hyperlinks
\usepackage{authblk}  % Used to override osajnl.cls author block style 

\usepackage[nonumberlist, nogroupskip, section=section, numberedsection=autolabel]{glossaries}
\glsdisablehyper
\newacronym{rl}{RL}{Reinforcement Learning}
\newacronym{umv}{UMV}{Ultra Mobility Vehicle}  
\newacronym{cmdp}{CMDP}{Constrained Markov Decision Process}
\newacronym{mdp}{MDP}{Markov Decision Process}
\newacronym{rsi}{RSI}{Reference State Initialization}
\newacronym{ppo}{PPO}{Proximal Policy Optimization}
\newacronym{mlp}{MLPs}{Multi-Layer Perceptrons}
\newacronym{rlhf}{RLHF}{Reinforcement Learning from Human Feedback}

\newcommand{\ig}{Isaac~Lab\xspace}
\newcommand{\fref}[1]{Fig.~\ref{#1}} % figures
\newcommand{\sit}[1]{\textit{\textbf{#1}}}
\usepackage{newfloat}
\DeclareFloatingEnvironment[
    fileext=lom,      % Extension for a list of movies (optional)
    listname={List of Movies}, % Title for list of movies (optional)
    name=Movie,       % The label used in the caption (e.g., Movie 1)
    placement=tbhp,   % Standard placement options
]{movie}
\begin{document} 

\begin{abstract} \bfseries \boldmath \sffamily
\vspace*{6pt}
Trials cyclists and mountain bike riders can hop, jump, balance, and drive on one or both wheels. This versatility allows them to achieve speed and energy-efficiency on smooth terrain and agility over rough terrain. Inspired by these athletes, we present the design and control of a robotic platform, \acrfull{umv}, which combines a bicycle and a reaction mass to move dynamically with minimal actuated degrees of freedom. We employ a simulation-driven design optimization process to synthesize a spatial linkage topology with a focus on vertical jump height and momentum-based balancing on a single wheel contact. Using a constrained \acrfull{rl} framework, we demonstrate zero-shot transfer of diverse athletic behaviors, including track-stands, jumps, wheelies, rear wheel hopping, and front flips. This 23.5 kg robot is capable of high speeds ($v_{max}$ = 8 m/s) and jumping on and over large obstacles (1 m tall, or 130\% of the robot's nominal height).
\vspace*{6pt}
\end{abstract}

\maketitle 

%%%%%%%%%%%%%%%%%%%%%%%%%%%%%%%%%%%%%%%%%%%%%%%%%%%
% FIGURE: Overview
%%%%%%%%%%%%%%%%%%%%%%%%%%%%%%%%%%%%%%%%%%%%%%%%%%%
\begin{figure*}[h]
    \centering
    \includegraphics[width=1.0\textwidth]{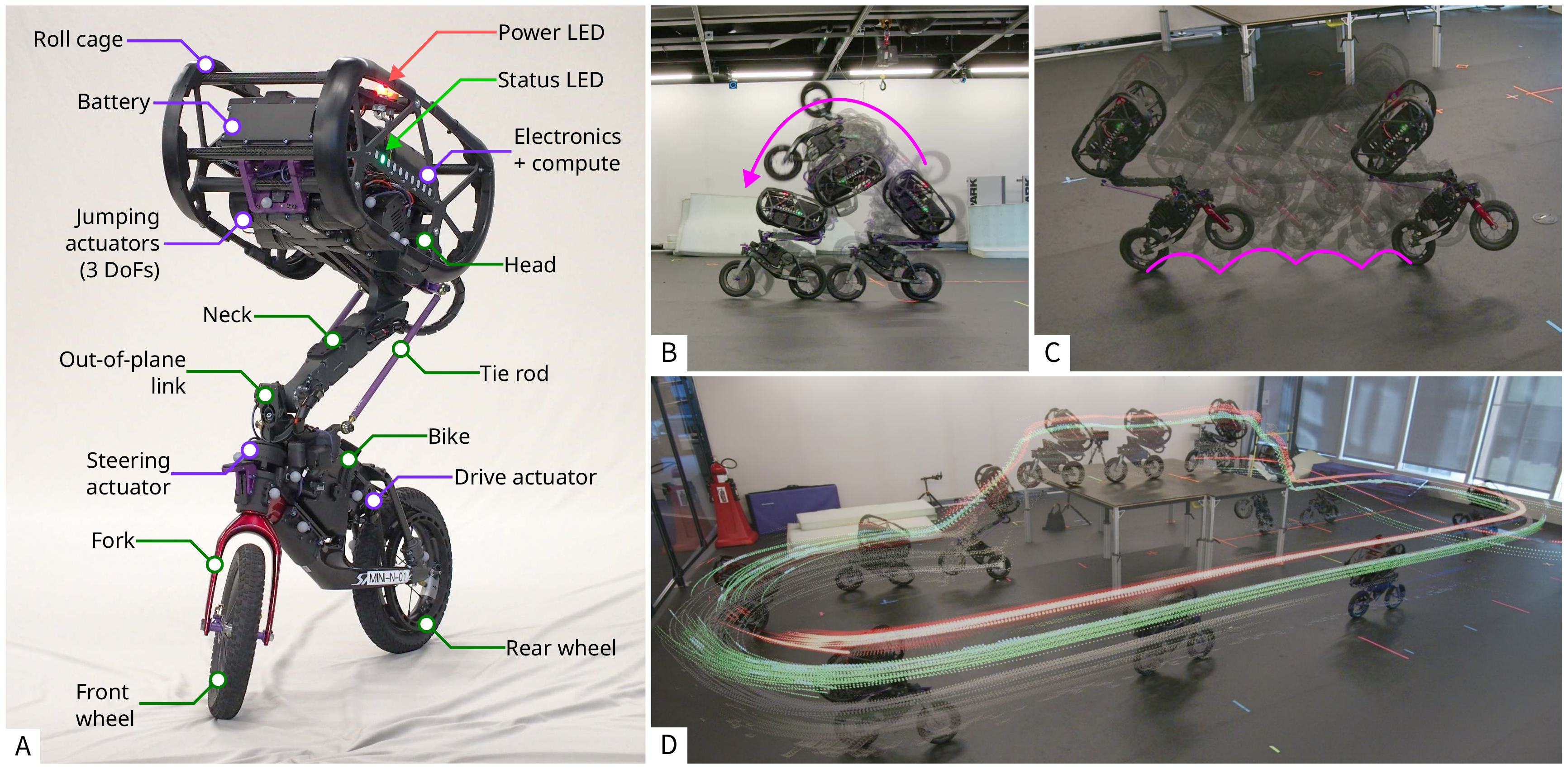}
    \caption{\textbf{The Ultra Mobility Vehicle (UMV) and its athletic repertoire.} 
    (\textbf{A}) Diagram of UMV, a bicycle-based robot with five actuated degrees of freedom [Links (green) and subcomponents (purple)]. It has steering and rear-wheel drive actuators for basic ground mobility. The design concentrates most of the robot's mass in the \sit{Head}, which connects to the \sit{Bike} through a spatial linkage.  This linkage consists of the \sit{Neck} and two tie rods. Powerful actuators in the \sit{Head} work through the linkage to let the robot ``throw'' its mass around, enabling dynamic behaviors. Composite images of (\textbf{B}) a front flip, demonstrating high pitch angular momentum and modulation of inertia via body tucking, (\textbf{C}) rear-wheel hopping, where the robot maintains balance like a single-legged hopper, (\textbf{D}) an autonomous table jump sequence, where the robot accelerates, vaults onto a 1-meter platform, traverses it, and lands stably.}
    \label{fig:collage}
\end{figure*}

\section*{Introduction}
Robots that navigate complex terrains, such as city streets and mountain trails, must be able to overcome a wide range of obstacles. Yet they are also expected to traverse long distances quickly and efficiently. While many robots are capable of either agile maneuvers or fast, reliable, and energy-efficient transportation, few perform well at both. Inspired by trials cyclists and mountain bike riders, who bridge this functionality gap, we introduce \gls{umv}, a dynamically stable robot that pairs bicycle-like locomotion with dynamic articulation of centroidal position and momentum to achieve multimodal locomotion with minimal degrees of freedom. \gls{umv} serves as a testbed to explore dynamic controls, state estimation, navigation, and the performance limits of acrobatic wheeled mobility.

\subsection*{Tradeoffs between Legs and Wheels}
Legged robots have demonstrated fantastic capabilities in recent years. By selecting discrete footholds, legged systems are able to handle vertical discontinuities that often pose challenges for wheeled platforms. For example, the humanoid Atlas~\cite{boston2021atlas} demonstrated the ability to jump onto boxes, bound between slanted surfaces, and even vault over obstacles. The quadruped ANYmal is able to climb over boxes that are taller than its standing height~\cite{hoeller2024anymal}. However, the leg morphologies that give these robots their impressive abilities are quite complex: Bipedal robots typically have six degrees of freedom (DoFs) per leg~\cite{chignoli2021humanoid} while quadrupeds tend to have three DoFs per leg~\cite{hutter2016anymal, bledt2018cheetah, johnson2013toward}. Therefore, both bipeds and quadrupeds typically have twelve actuated DoFs in the legs, and as DoF count increases, so does mass, manufacturing cost, computational complexity, and risk of mechanical failure. In contrast, wheeled robots can have as few as two actuated DoFs~\cite{getz1995control, tanaka2004self, goldfain2019autorally} and excel at fast, energy-efficient transportation on relatively smooth terrain, but have difficulty overcoming obstacles. Efforts to address these tradeoffs have given rise to the ``wheel-legged hybrid,'' where the feet of a legged robot are replaced with driven wheels~\cite{kashiri2019centauro, bjelonic2020rolling, unitree2025b2w}. Such robots have undeniable benefits; for example, Bjelonic and colleagues found that, compared to walking, wheels halved the cost of transport (CoT)~\cite{bjelonic2020rolling}. Going further, several works offset the cost of the additional wheel DoFs by removing leg DoFs. For instance, Ascento~\cite{klemm2019ascento} has just four actuated DoFs. Other examples with simplified leg mechanisms include Diablo~\cite{liu2024diablo} and Handle~\cite{boston2017handle}, both of which have demonstrated the ability to hop and traverse stairs. These wheel-legged hybrids use a multi-track wheel configuration (where wheels are arranged side-by-side) as opposed to a single-track wheel configuration (where wheels are arranged in tandem, like a bicycle or scooter). While multi-track systems are easier to stabilize, they require additional suspension mechanisms to handle roll and deflection caused by uneven terrain and dynamic motion. In addition, multi-track systems have an inherently wider footprint, and thus have more difficulty with tight spaces and small affordances. As such, they have greater mass, greater mechanical complexity, and reduced maneuverability.

Despite being more difficult to stabilize, the bicycle form has long been known to be versatile and practical. Elite trials cyclists demonstrate that it is possible to perform impressive acrobatic behaviors on a bicycle, including climbing over obstacles, leaping long horizontal and vertical distances from a standing position, traversing narrow beams, and balancing on a single contact patch~\cite{macaskill2023postcard}. Furthermore, elite mountain bikers demonstrate that the simplicity and durability of bicycles lend themselves well to handling steep drops and quickly traversing rugged terrain~\cite{teton2015unreal}. Therefore, we posit that a bicycle-based morphology has several advantages over multi-track systems in providing both agility and fast, efficient transportation.

\subsection*{Bicycles as Robots}
The bicycle has long served as a testbed for controls research. Early work on the linearized dynamics of bicycles~\cite{whipple1899stability, meijaard2007linearized} paved the way for approaches using steering input for stabilization~\cite{xiong2024steering, getz1995control}. Systems using steering-based strategies, such as Ghostrider~\cite{levandowski2010autonomous}, demonstrated high-speed field autonomy. At low speeds, relying solely on steering input for balance is challenging. Historically, this issue has been addressed via auxiliary stabilization mechanisms, such as reaction wheels~\cite{murata2025boy}, control moment gyroscopes~\cite{beznos1998control, zhang2014stationary}, or pendulums~\cite{seekhao2020development}. While effective for static stability, these additions increase system mass without contributing to propulsive power or agility. In contrast, Wang and colleagues recently demonstrated a track-stand with fork and rear wheel actuation only~\cite{wang2024equilibrium}, and recent simulation studies have explored momentum-based jumping~\cite{yuan2024dynamic}.

\subsection*{Reinforcement Learning for Agile Locomotion}
Recent advances in~\acrfull{rl} have made remarkable dynamic behaviors possible on wheeled and legged robots~\cite{hoeller2024anymal, hwangbo2019learning, miller2025high}. \gls{rl} provides robustness by simulating observation noise and randomizing physical parameters during training. Thus, model differences and state estimation uncertainties can be considered during training. Additionally, with \gls{rl} contact sequences with the environment do not have to be manually specified. It also adapts well to nonlinear systems and large state spaces, as demonstrated by record-breaking quadruped running speeds~\cite{li2025reinforcement, margolis2024rapid, miller2025high}, agile parkour maneuvers~\cite{cheng2024extreme, kim2025high}, and strong cross-terrain generalization~\cite{lee2020learning, he2025attention}. These methods are typically ``reference-free,'' relying on hand-engineered, multi-term reward functions. When designing rewards for behaviors becomes difficult, other works utilize motion imitation techniques from demonstration data, such as motion capture~\cite{peng2018deepmimic, peng2020learning}, videos~\cite{peng2018sfv, allshire2025visual}, or trajectories produced by optimization~\cite{fuchioka2022opt, liu2024opt2skill, miller2023reinforcement}. These methods have taken advantage of widely available motion capture datasets~\cite{mahmood2019amass, harvey2020robust} to accelerate progress in humanoid motion learning~\cite{he2025asap, chen2025gmt, he2024learning, he2024omnih2o}.

\subsection*{Learning Bicycle Stunts}
Early \gls{rl} studies showed that properly shaped rewards could achieve basic cycling behaviors~\cite{randlov1998learning}. More recent work has pushed the boundaries of what can be achieved via \gls{rl}. Hierarchical control frameworks have enabled bicycles to maintain stability and follow paths across irregular terrain~\cite{zhu2023deep}, while integrated planning and control systems allow for navigation through narrow spaces~\cite{zheng2023reinforcement}. Similar methods have been extended to humanoid robots performing scooter balancing~\cite{baltes2023deep}. Dynamic maneuvers such as ramp jumps have also received attention, with learning-based methods optimizing for takeoff and landing stability~\cite{zheng2022continuous}, and model-based analyses providing insights through inverse kinematics and Bayesian optimization~\cite{wang2024bayesian, wang2024attitude}. Tan and colleagues \cite{tan2014stunts} demonstrated the potential of \gls{rl} for learning a wide range of acrobatic stunts (including wheelies, bunny hops, and pivots), setting a benchmark for high-agility control.

\subsection*{Contributions}
This work presents a system design and control framework for \gls{umv}.
Our contributions include:

\begin{enumerate}
    \item \textbf{Novel Morphology for Athletic Robotic Motion:} We present a novel bicycle-based robot morphology capable of highly agile maneuvers, and we detail the principled system design process used to arrive at this morphology.
    We use computational models to develop a spatial linkage topology that both maximizes jump height and allows for effective dynamic balance on a single wheel.
    We engineer a lightweight and powerful hardware design capable of executing the desired dynamic motions.
    \item \textbf{Unified Learning Framework:} We introduce a unified learning pipeline built on a constrained RL formulation that synthesizes policies for a diverse set of athletic behaviors.
    We demonstrate that a single policy architecture, with minimal changes to the reward formulation, can master driving, balancing, wheelies, hopping, and flipping.
    These behaviors are crafted via careful modeling (``real-to-sim'') sparing the need for explicit datasets or demonstrations.
    \item \textbf{Experimental Validation:} We provide experimental validation of this hardware-software pairing, demonstrating performance on par with state-of-the-art legged systems.
    \gls{umv} achieves high-speed autonomous traversal ($v_{max}$ = 8 m/s) and demonstrates obstacle clearance by jumping onto platforms up to 1 meter in height (approximately 130\% of nominal height).
    Furthermore, we demonstrate the zero-shot transfer of acrobatic maneuvers, including continuous rear-wheel hopping and front flips, where the policy learns to tuck the body to accelerate rotation.
\end{enumerate}

Collectively, these results illustrate that by integrating agile, underactuated hardware with robust learning policies, we can combine the efficiency of wheeled platforms with the agility of legged systems, thereby expanding the performance envelope of dynamic robotic mobility.

A key element of \gls{umv}'s design lies in what is absent. We do not use reaction wheels, kick-stands, or stable polygons of support. Similarly, by executing lateral hops, the robot overcomes the non-holonomic constraints of wheels. In this regard, the robot follows the legacy of dynamic legged locomotion~\cite{Raibert_1986} by using compute to unlock its unique mechanical capabilities.

\begin{figure*}
	\centering
	\includegraphics[width=0.85\textwidth]{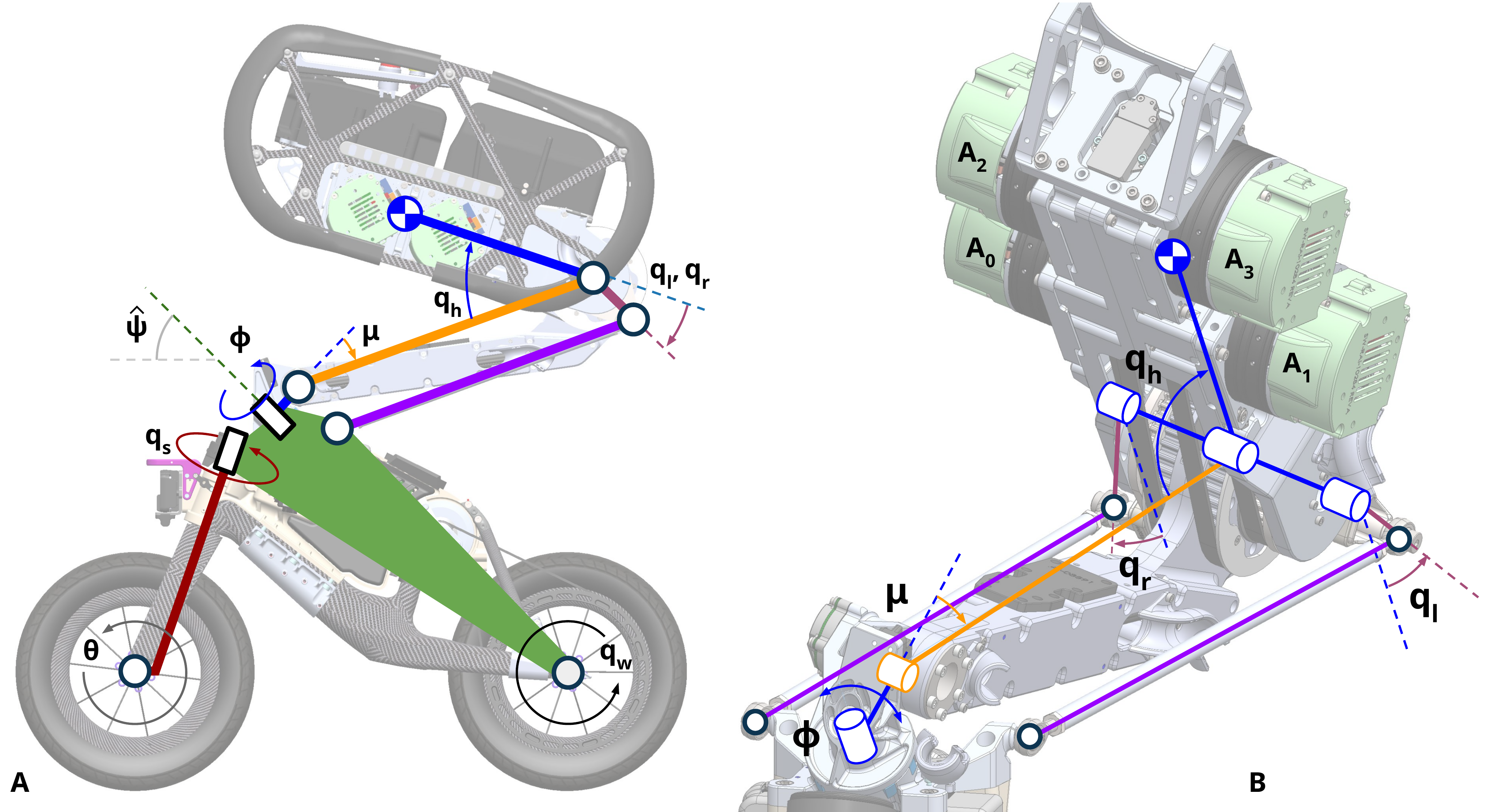}
	\caption{\textbf{Kinematic architecture and actuation strategy.} 
    (\textbf{A}) Side view showing joint layout. Joints $q_h$, $q_l$, and $q_r$ actuate the spatial linkage, allowing the \sit{Head} to shift its mass relative to the \sit{Bike} link. The design parameter $\hat{\psi}$ represents the fixed angle of the out-of-plane joint $\phi$ axis.
    (\textbf{B}) Detailed view of the spatial mechanism and actuator placement. Actuators $A_0$ through $A_3$ are remotized to the \sit{Head} to maximize the center of mass height and minimize leg inertia for jumping. $A_0$ and $A_1$ are mechanically coupled to drive the $\phi$ joint, providing lateral authority for single-wheel balancing.} \label{fig:kinematics}
\end{figure*}

In this paper, we present the design of the \gls{umv} system, its capabilities, and the control and state estimation methods employed on the platform to push the limits of robotic athleticism. We first present an overview of the hardware design~(Figs.~\ref{fig:collage} and~\ref{fig:kinematics}). Next, we showcase several experiments demonstrating \gls{umv}'s capabilities, including driving, turning in place and track-standing~(Movie~\ref{movie:one}), wheelies and hopping~(Fig.~\ref{fig:hops}), table jumps~(Fig.~\ref{fig:table_jump}), and flipping~(Fig.~\ref{fig:flip}). We then explain how the morphology of the robot was designed and optimized~(Figs.~\ref{fig:morphologies}, \ref{fig:jump_opt}, and \ref{fig:yoll_controllability_opt}). Finally, we cover details of the state estimation and \gls{rl} frameworks. The nomenclature used in this article is tabulated in Supplementary Table~\ref{tab:nomenclature}.

%%%%%%%%%%%%%%%%%%%%%%%%%%%%%%%%%%%%%%%%%%%%%%%%%%%
% MOVIE: RESULTS OVERVIEW 
%%%%%%%%%%%%%%%%%%%%%%%%%%%%%%%%%%%%%%%%%%%%%%%%%%%
\begin{movie*}[h]
    \centering
    \includegraphics[width=0.65\linewidth]{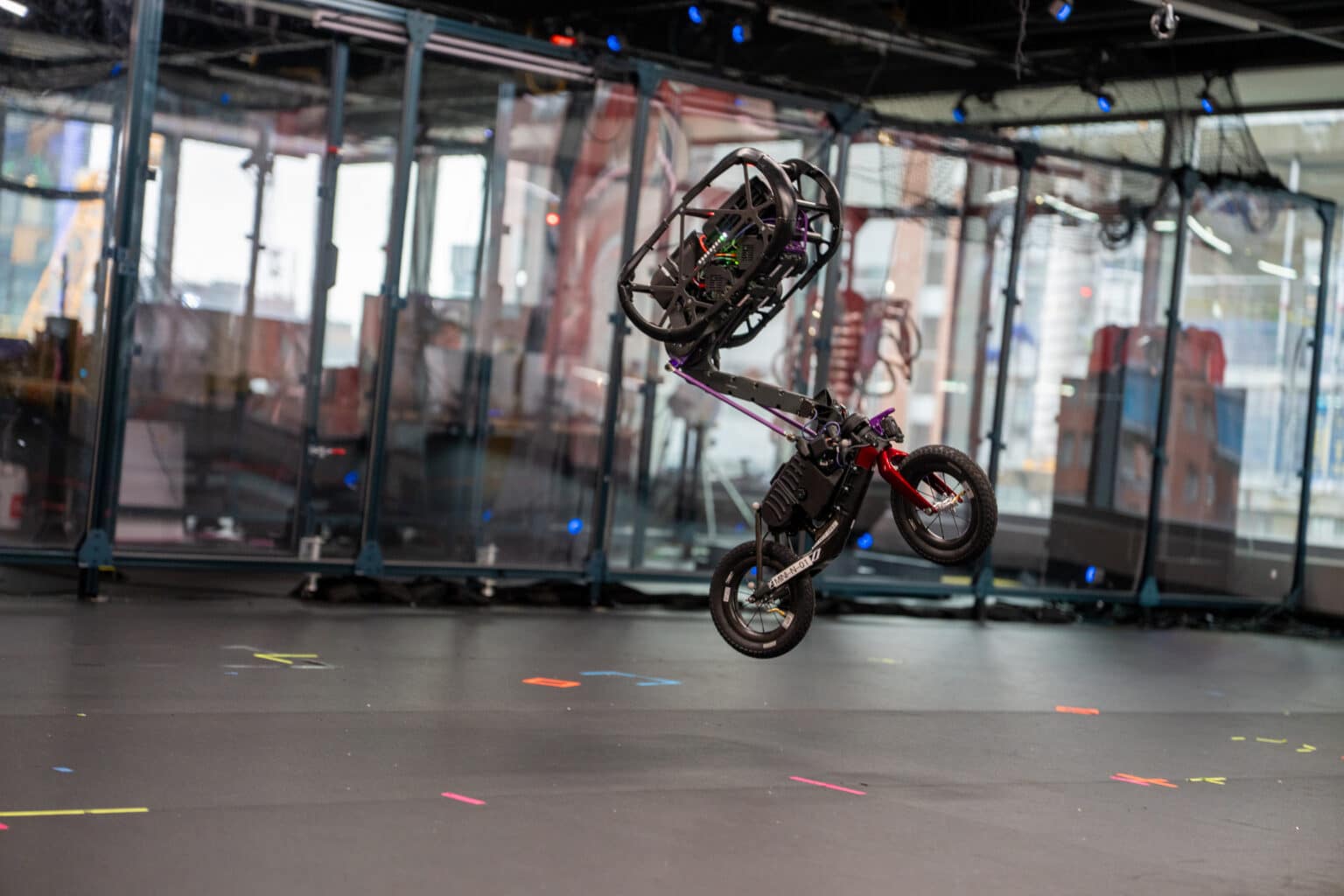}
    \caption{\textbf{\gls{umv} combines the bicycle form with legged agility,} allowing for athletic motions, including: (1) track-standing, in which the robot balances in place using only wheel and steering actuation; (2) backward driving, (3) ``shimmy-turning,'' in which the robot lifts the front wheel to yaw in place about the rear wheel, (4) lateral hopping on a single wheel, (5) front flipping, and (6) high-clearance table jumping.}
    \label{movie:one}
\end{movie*}

%%%%%%%%%%%%%%%%%%%%%%%%%%%%%%%%%%%%%%%%%%%%%%%%%%%
% FIGURE: Kinematics
%%%%%%%%%%%%%%%%%%%%%%%%%%%%%%%%%%%%%%%%%%%%%%%%%%%

\section*{Results}
\subsection*{System Overview}
The \gls{umv} robot, shown in~\fref{fig:collage}A, weighs 23.5 kg and is 0.8 m tall when fully crouched. It is composed of three main links, named the \sit{Head}, \sit{Neck}, and \sit{Bike} links. The robot has five actuated DoFs, denoted by the $q$ variables: [$q_h$, $q_l$, $q_r$, $q_s$, $q_w$]. Basic bicycle actuation comes through steering ($q_s$) and rear wheel drive ($q_w$) in the \sit{Bike} link. The three remaining DoFs, [$q_h$, $q_l$, $q_r$], are driven by actuators in the \sit{Head}. These actuators control the position and orientation of the \sit{Head} relative to the \sit{Bike} via a spatial linkage. This linkage consists of the \sit{Neck} link and left and right tie rod arms, and its kinematics are depicted in Figure~\ref{fig:kinematics}B. Actuators $A_2$ and $A_3$ are coupled to move $q_h$ directly, representing the angle between the \sit{Neck} and \sit{Head}. Actuators $A_0$ ($q_r$) and $A_1$ ($q_l$), coupled through the tie rods, work together to change the orientation of the \sit{Head} relative to the \sit{Bike}. As such, out-of-sagittal-plane motion of the \sit{Head} occurs when there is differential motion between $q_l$ and $q_r$. For convenience then, we define a conceptual ``serial'' topology that uses the angles $\phi$ and $\mu$ as an alternate representation of the kinematics. As shown below, this out-of-plane motion gives \gls{umv} unique balancing capabilities. As evidenced in Figure~\ref{fig:jump_opt}A, redistributing mass to higher links benefits jump height. Thus, the remotization of the jumping actuators, as well as the battery pack and compute, into the \sit{Head} is critical for maximizing jump height.
 
The \sit{Head} is composed of an aluminum 7075-T6 structural frame. It houses the jumping actuators, as well as compute, batteries, and most of the robot's electronics. These are protected by a roll cage of carbon-fiber reinforced polymer (CFRP) plates and tubes, with an outer layer of foam to damp impacts. The \sit{Neck} is machined from 7075-T6 in two halves. Like the \sit{Head}, its ``clamshell'' structure maximizes cross-sectional second moment of area per unit mass, improving stiffness against lengthwise torsion and lateral forces. An additional benefit is that electrical cabling is routed through the hollow center. The \sit{Bike} link consists of a magnesium AZ31B-H24 structural element clamped over the down tube of a children's CFRP push bike (specifically, a Specialized Hotwalk Carbon). The magnesium structure houses the steering and rear wheel drive actuators as well as the communication and power distribution electronics. The Hotwalk's CFRP fork and wheel rims are also used, where the output pulley of the rear drive belt is bonded to the rear rim. Additional struts reinforce the rear stays to prevent deflection from loosening the rear wheel belt drive during takeoff and landing.

\subsection*{Electronics}
\gls{umv}'s formidable power draw requirements during jumping are a focal point of the robot's electrical design. For example, jumping onto a 1 meter table results in a power draw spike with a peak of 4.5 kW over a 0.25 second period, consequently undergoing a voltage sag of up to 20 V (See~\fref{fig:table_jump}). \gls{umv}'s power source is a lithium-ion battery pack providing a maximum voltage of 58.8 V. The battery pack is protected by a BMS with timed current limit thresholds. The original firmware limits were modified to increase maximum discharge (within safe limits) at the expense of cycle life. These changes increased the maximum allowable power draw from 4640 W for 23 ms to 7356 W for 80 ms. Furthermore, regeneration clamping is used to prevent back-EMF spikes from damaging the electronics and batteries upstream of the motors.

The robot's main computer is an UP Xtreme i12, which runs a 1 kHz control loop responsible for reading from and writing to the connected devices (motor drivers, IMUs, encoders, etc.), as shown in Fig.~\ref{fig:electronics}.
By having three 3-DoF acceleration sensors on each rigid link, we are able to measure the instantaneous linear acceleration, angular acceleration, and angular velocity of each link.
(Although this could have been achieved with two IMUs and four accelerometers, reuse of the IMU saved us development time and complexity.)
This allows the estimation of front and rear ground contact state via a two-body lumped dynamics model that estimates tire contact forces and classifies probable wheel contacts.

\subsection*{Actuators}
Actuators $A_0$ through $A_3$ (see Fig.~\ref{fig:kinematics}B) are each composed of a permanent magnet synchronous motor (PMSM) driving a single-stage planetary transmission.
The actuators then drive the joints of the spatial linkage through stiff timing belts.
Each motor controller (ODrive Pro) closes a PD loop on position and velocity setpoints from the robot's main computer at 8 kHz.
As mentioned above, actuators $A_2$ and $A_3$ are coupled together to drive a single actuated DoF, $q_h$.
This arrangement lets us meet the torque and power requirements of the $q_h$ axis while keeping the link's CoM close to the sagittal plane.
For both the rear wheel drive $A_4$ and steering $A_5$ actuators, the transmission between motor and output is a single belt drive.
For belt tensioning, while the rear drive uses an idler, the positions of the other actuators can be adjusted.
By avoiding the use of idlers where possible, we extend belt life and improve ease of assembly and maintenance.
More details on actuator specifications can be found in Supplemental Table~\ref{tab:actuators}.

\subsection*{Experiments}
We developed and tested a wide range of agile and mobile capabilities on \gls{umv} using \gls{rl}.
Our learned policies generalized zero‐shot from simulation to on-robot in a lab environment.
The behaviors covered in this work include driving, wheelies, rear-wheel hopping, front flips, and jumping.
Although each behavior required a different policy, they all used a similar \gls{rl} pipeline.

\subsubsection*{Driving and Track-Standing}
In this experiment, we tested the robot's ability to drive and balance dynamically.
The driving behavior tracks the forward velocity and heading commands (coming from either a joystick or higher-level planner) as the robot moves across flat ground.
While bicycles are known to have some passive stability while moving forward, backward movement is inherently unstable~\cite{meijaard2007linearized}.
Movie~\ref{movie:one} shows, though, that a learned policy allows \gls{umv} to maintain its balance driving both forwards and backwards.
The robot is also able to turn in place about a (stationary) rear wheel contact point.
We call this the ``shimmy-turn'', where the robot uses the reaction mass of the \sit{Head} to quickly step the front wheel laterally, creating yaw about the rear wheel.
Rapidly repeating this pattern executes in place turning. This behavior is differentiated from small radius turns with both wheels in ground contact, which requires a small amount of forward motion.
The ``shimmy-turn'' behavior emerged from the learned policy as a viable control strategy without being explicitly designed.
Finally, \gls{umv} is capable of holding a ``track-stand.'' This behavior is basically balancing in place, over the contact line between the front and rear wheels, with very little motion.
It can be performed using just fork and rear wheel actuation, while adding \sit{Head} actuation further improves balance.

\subsubsection*{Wheelies and Hopping}
\begin{figure}[htb] 
	\centering
	\includegraphics[width=1.\columnwidth]{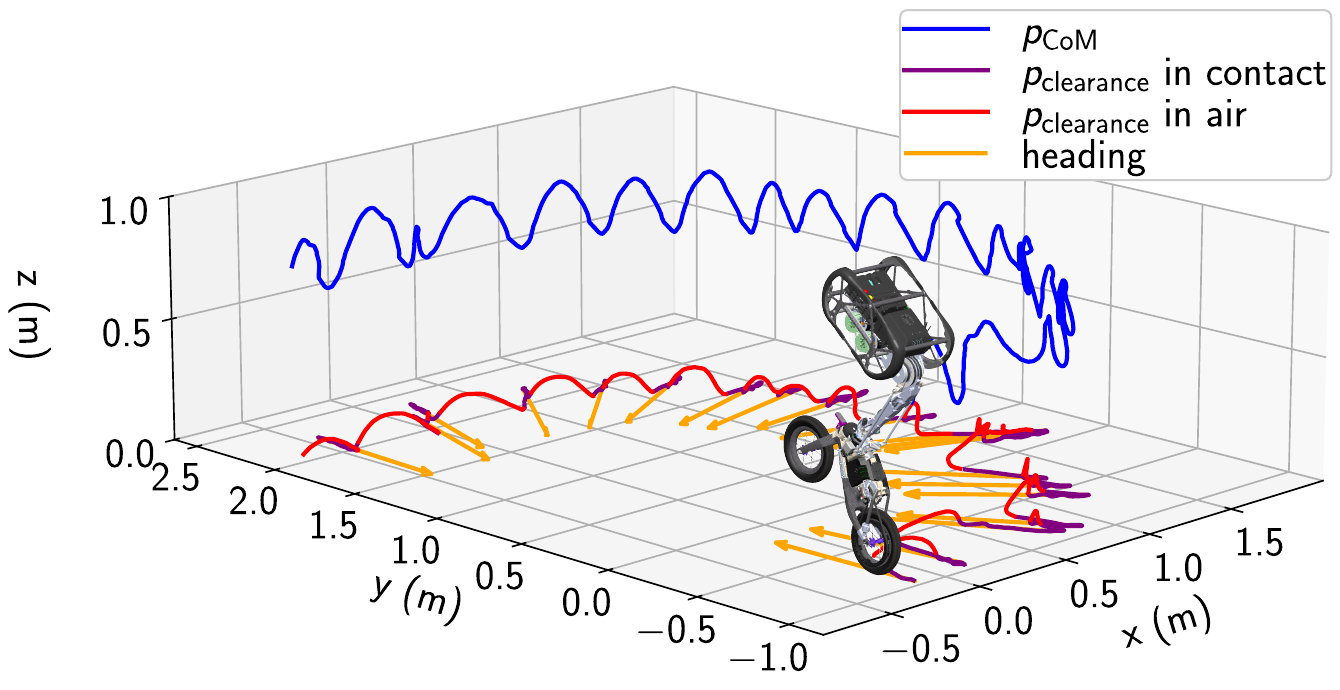}
	\caption{\textbf{Positions, footstep locations, and headings over a roughly 12-second period of lateral hopping on hardware.}  $p_{\text{CoM}}$ shows the oscillation of the whole-body center of mass throughout the hopping motion.
    The traces at ground level ($p_{\text{clearance}}$) indicate the lowest point of the rear wheel in both flight and stance.
    These visualize a learned stabilization strategy where the rear wheel rolls backward and forward during the stance phase to maintain balance, analogous to a unicycle.}
	\label{fig:hops}
\end{figure}

Using a single policy, \gls{umv} is able to do both continuous wheelies and rear-wheel hopping.
The goal of these behaviors is to track desired forward velocity and heading commands while keeping the front wheel off the ground.
In the hopping mode, the robot is also rewarded for tracking lateral velocity commands.
To encourage hopping, we use a predefined periodic contact sequence.
The policy also controls the duration of each contact phase, which improves its robustness under disturbances.
Hopping allows the robot to locomote monopodally; it enables movement in arbitrary directions, a useful ability in constrained spaces.
Figure~\ref{fig:hops} depicts a rear wheel hopping experiment where the robot was commanded to laterally hop in a semicircular pattern over a period of 12 seconds.
During hopping, it can be observed that the robot rolls its rear wheel backward during the contact phase before rolling it forward again for lift-off.
This behavior is likely a stabilizing adaptation by the policy, as it is similar to the forward-backward movement required to stabilize a wheelie.
We also found that extension of the articulated mechanism is critical for stabilization, as it reduces the tipping frequency.

\begin{figure*}
	\centering
	\includegraphics[width=0.95\textwidth]{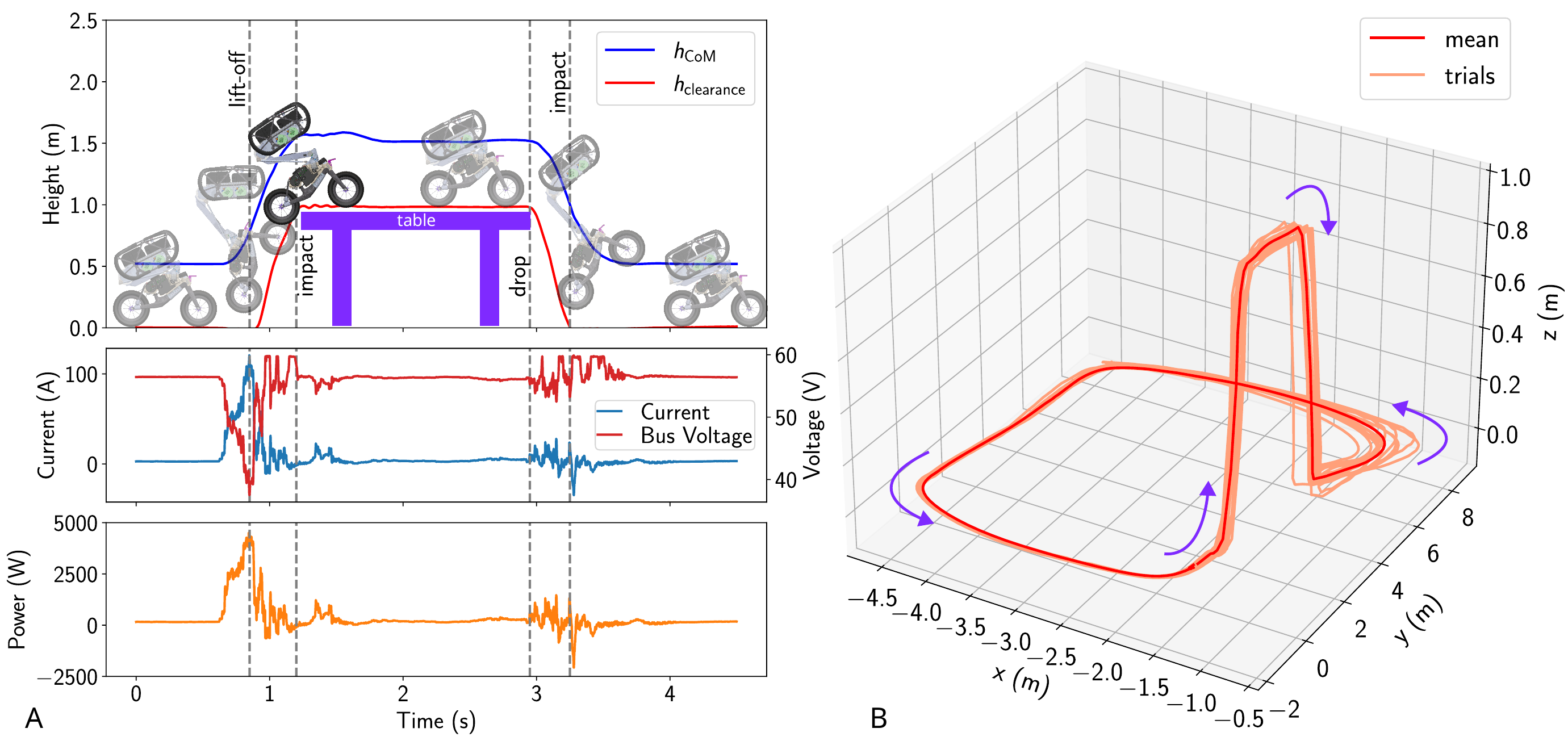}
	\caption{\textbf{Energetics and repeatability of high-clearance jumping.} 
    (\textbf{A}) System performance during a 1-meter table jump.
    The top plot shows the CoM height ($h_{\text{CoM}}$) and rear wheel clearance ($h_{\text{clearance}}$).
    The vertical dashed lines mark the high-impulse lift-off and landing impact events. The bottom plots illustrate the electrical demand;
    the system draws peak power (approx. 4.5 kW) immediately prior to lift-off, causing a significant voltage sag, while landing impacts briefly drive the actuators into regeneration (negative power).
    (\textbf{B}) Spatial repeatability of the rear wheel trajectory ($p_{\text{clearance}}$) across 15 consecutive autonomous jumps.
    The tight clustering of trajectories during the ascent phase demonstrates the precision of the learned policy, with increased variance occurring only post-impact during the drop-landing.}
	\label{fig:table_jump}
\end{figure*}

\subsubsection*{Table Jumps}
The table-jumping experiment tests \gls{umv}'s ability to autonomously position itself in front of a table, leap onto and drive across it, drop off the other side, and repeat.
This experiment demonstrates a number of key capabilities of \gls{umv}.
First, it involves the combination of multiple higher-level dynamic behaviors: driving, following waypoints, jumping onto a table, and jumping down from a table.
Secondly, it tests the extremes of \gls{umv}'s physical performance requirements.
Figure~\ref{fig:table_jump}A depicts the robot's change in height as it leaps onto a table, as well as the electrical demands on the system.
Power draw reaches its peak and, consequently, bus voltage reaches its lowest point just before lift-off.
When the robot drops off the table, ground impact causes a voltage spike while power flow becomes regenerative;
the actuators for the spatial linkage, which support the \sit{Head}, briefly become generators.

Finally, this experiment displays the system's repeatability.
Figure~\ref{fig:table_jump}B tracks the position of the lowest point of the rear wheel across fifteen consecutive jumps, showing \gls{umv}'s consistency and robustness.
Greatest variability occurs immediately after dropping off the table, as ground impacts present difficulties for both the sim-to-real gap and state estimation.

\begin{figure}[htb]
	\centering
	\includegraphics[width=1.\columnwidth]{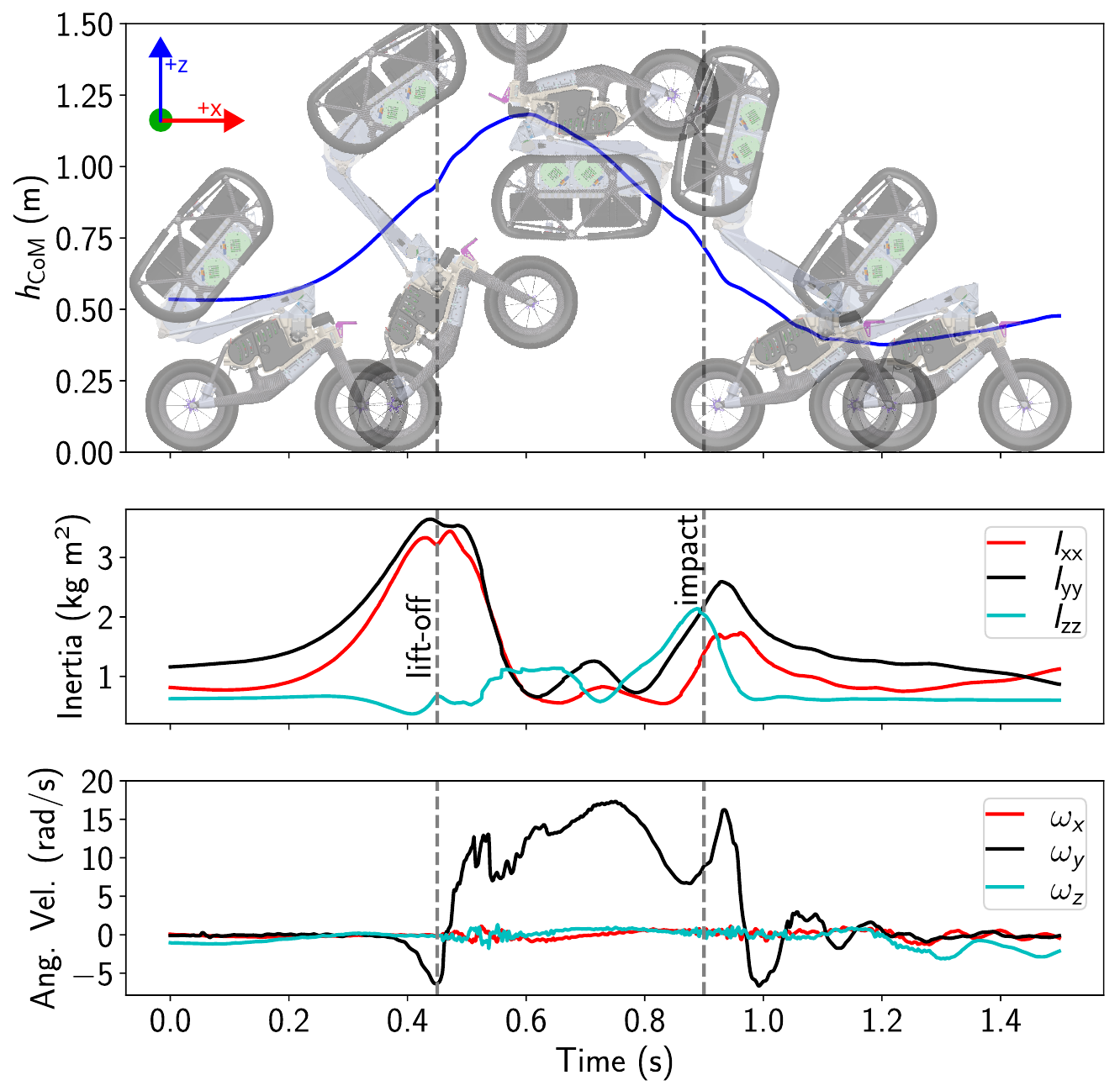}
	\caption{\textbf{Angular momentum conservation during a front flip.} 
    This data trace illustrates the physics of the flip maneuver shown in Fig. 1B.
    Following lift-off (first vertical dashed line), the robot actively tucks its body, drastically reducing the pitch inertia $I_{yy}$ (middle plot, black line).
    This reduction in inertia facilitates a rapid increase in pitch angular velocity $\omega_y$ (bottom plot, black line), peaking at approx.
    17 rad/s, allowing the robot to complete the rotation before extending for impact (second vertical dashed line).
    Shown in the bottom plot is the angular velocity of the \sit{Bike} link.}
	\label{fig:flip}
\end{figure}

\subsubsection*{Front Flips}
The purpose of the front flip experiment (seen in Fig.~\ref{fig:collage} and Movie~\ref{movie:one}) is to demonstrate \gls{umv}'s ability to perform stylistic dynamic maneuvers.
During a front flip, the robot attempts to move forward, jump, tuck into a forward spin, and land in a stable manner.
Figure~\ref{fig:flip} shows that after lift-off, the robot tucks to reduce its centroidal moment of inertia about the pitch axis.
This movement corresponds with a dramatic increase in pitch angular velocity, peaking at approximately 17 rad/s, before the process reverses.
Compared to a table jump, the contortion required for a flip allows for less time and space with which to cushion the robot's landing.
As a result, the robot hits the ground with considerable force.
Despite the mechanical, control, and state estimation difficulties this impact would be expected to present, the robot is able to land without running into joint limits and stably transition into a driving behavior (See Movie~\ref{movie:one}).

\section*{Discussion}

We have presented the system design and control of the \gls{umv} robot, a dynamically stable acrobatic wheeled robot that explores the perspective that mechanical underactuation, when coupled with high-bandwidth learning-based control, can yield athletic robotic motion.
By integrating a bicycle chassis with a powerfully actuated reaction mass, we demonstrated that static stability mechanisms are not required for versatile mobility.
Instead, \gls{umv} treats stability as a dynamic control task, leveraging the conservation of angular momentum and centroidal manipulation to achieve behaviors such as track-standing, rear-wheel hopping, and front flips.
Results show that constrained reinforcement learning can effectively navigate the complex, non-minimum phase dynamics of this morphology.
It also discovered emergent solution strategies, such as the zero-velocity ``shimmy-turn,'' which is difficult to manually encode in classical control templates.
The successful zero-shot transfer of these maneuvers suggests that the intersection of computational design and data-driven control is a pathway for developing the next generation of agile locomotion.

The \gls{umv} robot bridges the gap between the energetic efficiency of wheeled platforms and the obstacle negotiation capabilities of legged systems. While wheel-legged hybrids have been explored previously~\cite{boston2017handle, klemm2019ascento, liu2024diablo}, \gls{umv} distinguishes itself by being a low DoF, single-track bicycle-based robot. Demonstrating field speeds up to \unit[8]{m/s}, \gls{umv} outperforms the steady-state cruising speeds of standard quadrupedal platforms~\cite{hutter2016anymal, KAIST_HOUND} and demonstrates performance on par with that of wheel-legged robots such as ANYmal on Wheels (max speed of 10 m/s)~\cite{hoeller2024anymal} and the Unitree B2-W (max speed of \unit[6]{m/s})~\cite{unitree2025b2w}. \gls{umv}'s combination of speed and small spatial footprint could help with applications such as urban courier service, or long-distance reconnaissance over rough terrain. \gls{umv}'s capabilities, coupled with its relatively few actuated degrees of freedom, put it in a favorable position compared to quadrupedal or bipedal robots, offering potential advantages regarding manufacturing complexity, weight reduction, and maintenance.

The eventual goal for \gls{umv} is to enable high-speed autonomous traversal of varied environments without off-robot sensing (see also Supplementary Movie \ref{movie:field}), which necessitates a number of improvements to our hardware, state estimation, perception, and control systems. Regarding \gls{umv}'s hardware, the robot could be ruggedized against water, dust, and intrusion. In addition, we are developing custom motor drivers that will improve our low-level joint control, and batteries with greater power density. We are also investigating a self-righting functionality that does not require additional DoFs. The results documented in this work are from a laboratory setting with a calibrated high-fidelity motion capture system that provides ground-truth and isolates reported values from state-estimation bias and drift. We are currently testing LiDAR-inertial odometry and learned state estimation instead of using motion-capture. Part of this effort involves improving the automation of reference generation to expand \gls{umv}'s repertoire. Finally, the sim-to-real gap remains a challenge for dynamic, contact-rich behaviors. We are employing generative simulators for richer training distributions beyond static domain randomization, which supports policies that move beyond heavily-shaped rewards.

\section*{Materials and Methods}\label{methods}

\subsection*{Design Optimization for Agility}
The objective of this study was to empirically discover an optimal robot for execution of high-agility maneuvers. To this end, our engineering design process was heavily guided by simulation. However, a single optimization objective was difficult to set because we intended \gls{umv} to be adaptable to a wide variety of athletic tasks. Vertical jumping ability is important to many of \gls{umv}'s tasks. It is also one of the most physically demanding behaviors for a robot in terms of actuator requirements, mass distribution, and power draw. For these reasons, we chose maximization of jump height as the foremost consideration in the design process. We primarily measured jump height based on whole-body CoM location ($\max(h_\text{CoM})$), but clearance height ($\max(h_\text{clearance})$)---the height that the lowest point of the rear wheel can clear---was used in cases where scale was an optimization variable, as being tall can provide a benefit to $\max(h_\text{CoM})$ without improving a robot's ability to clear obstacles (for example, it may be unable to tuck in time). A secondary measure we looked at was ``contact ratio'' ($c$), defined as the proportion of time spent in ground contact during a jump, from full crouch to apogee. A smaller $c$ implies a higher jump frequency, which is desirable.
\begin{equation}
c = \frac{t_\text{lift-off} - t_0}{t_\text{apogee} - t_0}
\end{equation}

Despite our focus on maximization of jump height, it should be noted that we eschewed mechanisms specialized for jumping in favor of more general mechanisms which can be optimized for jumping. For example, parallel springs~\cite{klemm2019ascento}, serial springs~\cite{haldane2016robotic}, and latched spring mechanisms~\cite{hawkes2022engineered} are known to be highly beneficial for jump height. However, spring mechanisms tend to enhance one behavior at the expense of most others. On the other hand, sole use of electrical energy storage, through the use of batteries and motors, can lead to a more versatile robot.

\begin{figure*}
    \centering
    \includegraphics[width=1.\textwidth]{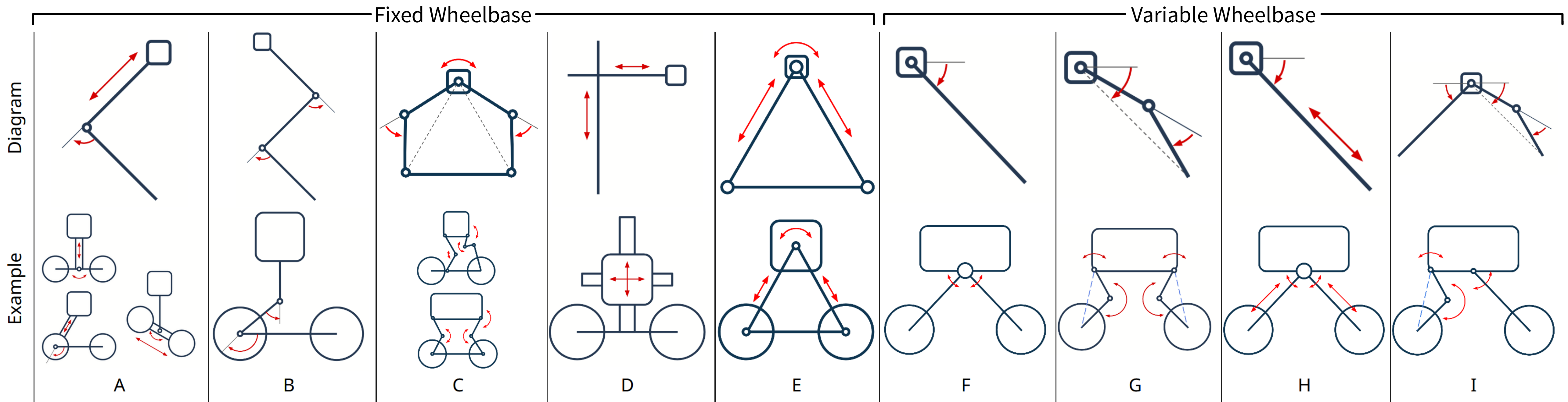}
    \caption{\textbf{Design space exploration for jumping morphology.} 
    A qualitative overview of the kinematic topologies evaluated during the design phase.
    Morphologies were assessed based on their theoretical capacity for vertical jumping, including the ratio of extended height to crouched height per unit of link length.
    Concepts ranged from simple prismatic legs (A) to complex multi-linkage systems (H, I).
    Topology (B) was selected for \gls{umv} as it offered the optimal trade-off between stroke length, mass distribution, and mechanical complexity.}
    \label{fig:morphologies}
\end{figure*}

Figure~\ref{fig:morphologies} depicts a broad set of morphologies that we evaluated during the design process. Initial down-selection was qualitative. For example, we ruled out morphologies that use prismatic joints due to harsh tradeoffs between workspace and mass. We then examined the sensitivity of clearance height to link lengths for each simulated morphology. The relationship between link length, mass, and moment of inertia was determined by approximating links as cylinders. All morphology comparisons were done at the same nominal total mass. All simulated morphologies used the same jumping actuators (CubeMars AK10-9), which we selected via a systematic comparison of commercial off-the-shelf actuators on key metrics such as peak power density, torque density, and torque magnitude.

\begin{figure*}
    \centering
    \includegraphics[width=1.0\textwidth]{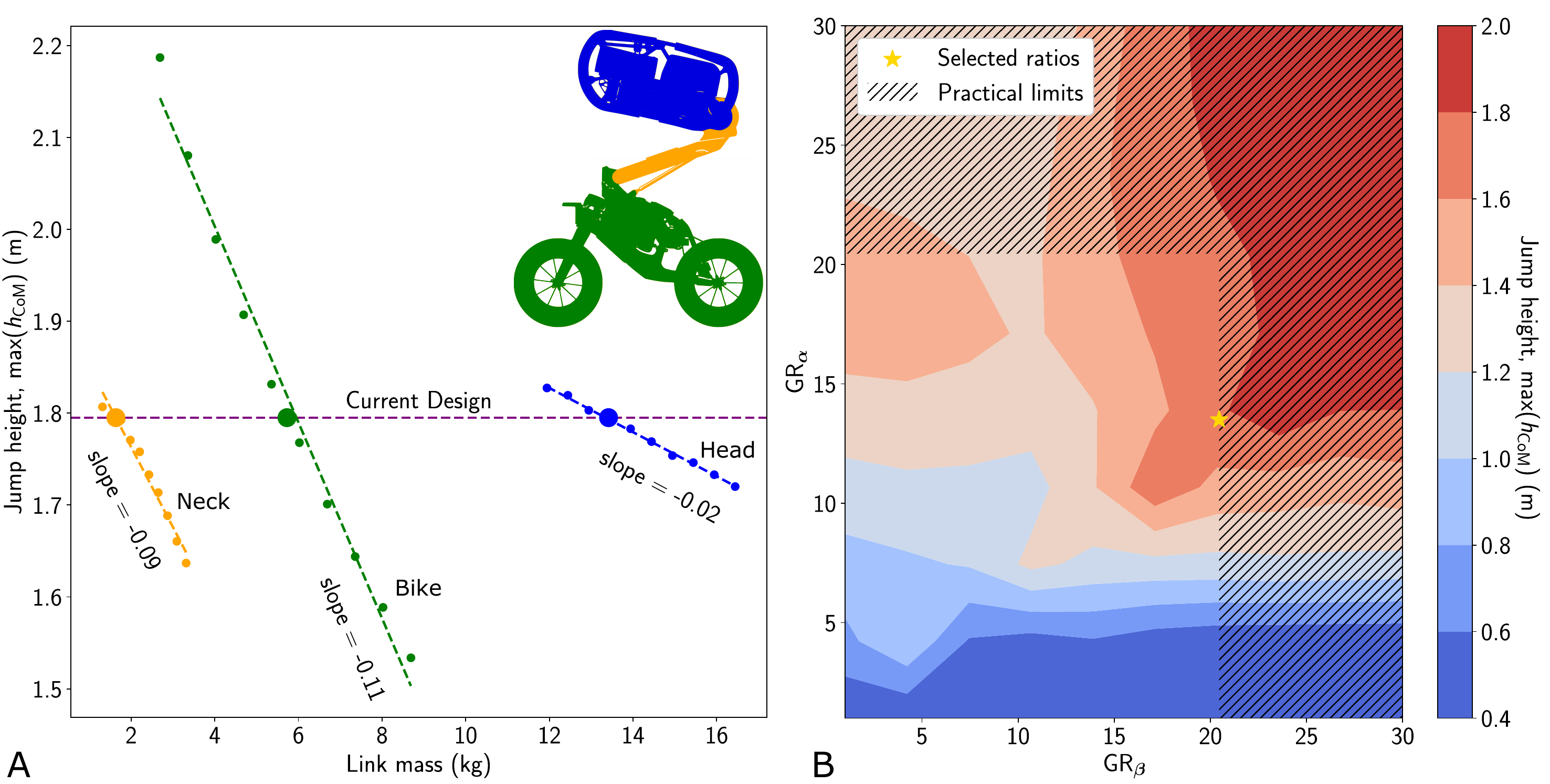}
    \caption{\textbf{Mass distribution sensitivity and optimization of transmission ratios.}
    (\textbf{A}) Sensitivity analysis of link mass on jump height. We varied each link mass while keeping the other masses at their nominal value (``Current Design''). The slope for the \sit{Head} (blue) is shallower than the slopes for the \sit{Bike} (green) and \sit{Neck} (yellow), which indicates that remotizing mass to the top of the robot has a smaller penalty on jumping performance. The current design is already the result of a concerted effort to minimize the mass of the lower links within practical limits.
    (\textbf{B}) Gear ratio optimization landscape. The heat map displays jump height as a function of the gear ratios for actuators $A_0$ and $A_1$ ($GR_{\alpha}$) and actuators $A_2$ and $A_3$ ($GR_{\beta}$). 
    The star indicates the selected ratios, chosen to maximize performance while remaining within the ``Practical limits'' (hatched area) imposed by pulley diameter constraints and gearbox complexity.}
    \label{fig:jump_opt}
\end{figure*}

The morphology depicted in Fig.~\ref{fig:morphologies}B exhibited particular promise due to a high ratio of extended height to crouched height per unit link length. We therefore performed further investigations on this morphology. First, we needed to select the robot's gross size. To do this, we compared tradeoffs between clearance height and contact ratio as functions of length scale, and factored in the possibility of using coupled actuators for joint $q_h$ to improve jump height despite the increased mass. As scale increases, jump height increases with diminishing returns until it plateaus, whereas contact ratio increases at a linear rate until the maximum value of 1 is reached (See Supplementary Fig.~\ref{fig:coupled}). Therefore, a balance must be struck between the desire to increase jump height and the desire to decrease contact ratio. Balancing these tradeoffs and taking practical considerations into account, such as off-the-shelf electronics mounting and lab space limitations, we settled on a scale equating to a \sit{Neck} length of 0.4 m. Secondly, because jump height sensitivity to mass changes is far lower for the \sit{Head} (as depicted in Fig.~\ref{fig:jump_opt}A), we determined that actuators should be remotized to the \sit{Head} with the use of a spatial linkage. Having settled on a detailed kinematic layout and nominal dimensions, we developed a first draft of the robot's design in CAD. We then updated our simulation with an estimate of the robot's mass properties extracted from the CAD model. Finally, we performed a gear ratio optimization~(Fig.~\ref{fig:jump_opt}B). Although further performance gains could be attained with slightly larger gear ratios, the optimization problem was constrained by practical limits to pulley diameters while avoiding increased complexity, reflected inertia, impact fragility, friction, backlash, and other deviations from the proprioceptive actuator paradigm~\cite{wensing2017proprioceptive}.

\subsection*{Design Optimization for Lateral Balance On One Wheel}
In addition to vertical jumping, robust balancing also remains critical. It is well established that a bicycle in motion may balance with fork actuation only~\cite{hand1988}. On the other hand, balancing on a single wheel requires momentum control outside of the sagittal plane. Several unicycle robots have addressed this challenge with the use of either reaction wheels~\cite{Vos1992, Geist2022} or control moment gyroscopes~\cite{Brown1996}. However, flywheels would add a significant mass penalty to our system without contributing to the linear impulses required for jumping. In contrast, a human cyclist is able to balance on a single wheel through effective lateral motion~\cite{Okawa2009}. Given a morphology that is already optimized for vertical jumping (as discussed above), we looked at whether adding an extra DoF for out-of-plane motion improves the system's ability to continuously balance on the rear wheel (i.e., a continuous wheelie). And if so, how should this additional DoF be placed within the morphology? To answer these questions and thereby optimize our design, we use an aggregate measure of controllability to quantify the ease of control in balancing on a single wheel. Controllability denotes the ability of a system to be driven within its configuration space using only admissible inputs. 
More formally,
\begin{equation}
\begin{split}
\boldsymbol{\Xi}(\mathcal{S}) &= \int\limits_{\boldsymbol{x}_s,\boldsymbol{x}_f\in \mathcal{S}} \boldsymbol{E}(\boldsymbol{x}_s, \boldsymbol{x}_f)d\mathcal{S}\\
\text{where}\;\;\boldsymbol{E}(\boldsymbol{x}_x, &\boldsymbol{x}_f) = \min \int_0^{\infty}\boldsymbol{u}(t)^T\boldsymbol{u}(t)dt\\\text{s.t.}\;\boldsymbol{x}(0)&=\boldsymbol{x}_s\;\text{and}\;
\boldsymbol{x}(\infty)=\boldsymbol{x}_f\end{split}
\label{eq:lticontrollability}
\end{equation}
where $\boldsymbol{u}$ are the control inputs and $\boldsymbol{x}$ the state variables describing the system. Depending on the underlying dynamics, it may not be possible to drive the system between particular $\boldsymbol{x}_s$ and $\boldsymbol{x}_f$, and the effort $\boldsymbol{E}$ in such a case goes to $\infty$. $\boldsymbol{\Xi}(\mathcal{S})$ is an accumulation of the minimum required control effort to drive within the region $\mathcal{S}$. A more controllable system will have a lower $\boldsymbol{\Xi}(\mathcal{S})$, indicating a lower required effort to regulate the system within $\mathcal{S}$. Note that although the above definition is quite general, computing it for any system with nonlinear dynamics, while accounting for input limits and other constraints, is non-trivial and requires numerical trajectory optimization methods. For linear time invariant (LTI) systems, closed form, computationally cheaper solutions exist~\cite{lindmark2018}. As approximations of the minimum control effort for nonlinear systems, we use local estimates based on linearizations of the system dynamics about several different configurations and aggregate them together as follows.
\begin{equation}
\underset{\mathcal{S}=\text{Conv}(\mathcal{S}_i)}{\boldsymbol{\Xi}(\mathcal{S})}\approx\sum_i\boldsymbol{\Xi}_{\text{local}}(\mathcal{S}_i)
\label{eq:aggregate_controllability}
\end{equation}
We evaluate this aggregate measure of controllability for continuous balancing on the rear wheel by sweeping over several statically stable configurations of wheelies, and computing local estimates (\eqref{eq:lticontrollability}) from linearizations of the dynamics~\cite{lindmark2018}. We confirmed that adding an out-of-plane joint ($\phi$ as shown in \fref{fig:kinematics}) between the Bike link and joint $\mu$ significantly improved balancing performance. We then determined the fixed orientation $\hat{\psi}$ of this axis that minimized required control effort~(Supplementary \fref{fig:yoll_controllability_opt}A). Also, we found that adding another DoF in addition to $\phi$ provides diminishing returns~(\fref{fig:yoll_controllability_opt}B). All of these observations led us to settle on the morphology depicted in Fig.~\ref{fig:kinematics}.

\subsection*{State Estimation}
In order to control \gls{umv}, an estimate of all states and their first time derivatives is needed. This complete state is comprised of the 6-DoF floating base state and the six serial joint states. Because the front wheel's mass is low, its velocity and acceleration can generally be neglected. Of the six serial joint states [$q_h$, $\mu$, $\phi$, $q_s$, $q_w$, $\theta$], the position and velocity of the five that are necessary for dynamics are directly measured (i.e., all but $\theta$). The largest state estimation challenge required to drive \gls{umv} to the edge of its capability is estimation of the 6-DoF floating base state and its derivative. This state is comprised of the 3-DoF orientation, parameterized in our optimization as a 3$\times$3 rotation matrix, and the 3-DoF position of the \sit{Bike} link frame. The linear velocity (position derivative) must be estimated, as it cannot be directly observed. The angular velocity is directly measured via rate gyro from an IMU and does not require explicit estimation. As an engineering compromise, and to let us focus on platform design and control, we utilize a motion capture system to aid in state estimation while in the lab. Motion capture estimates are fused with a high rate onboard IMU to produce quality base-state estimates. The robot does support self-contained state-estimation, which has been demonstrated in field operations. Retro-reflective markers on the robot are tracked by an off-the-shelf Optitrack motion capture system to produce position and orientation estimates at 120Hz. In order to provide high rate, low latency estimation to controllers, we use the GTSAM~\cite{gtsam} factor graph optimization package to fuse the 1kHz IMU signal and the lower rate Optitrack signal. We adopted a bipartite factor graph architecture~\cite{goldfain2019autorally} where the IMU angular rates and linear accelerations are integrated at 1kHz to provide a smooth low latency pose estimate, and the Optitrack measurements are decimated to 30Hz and used to correct this high frequency estimate. A factor graph tracks position, orientation, linear velocity, and gyro and accelerometer biases. We also use both position and orientation measurements from the motion capture system. Because our system is highly dynamic and sometimes generates significant noise or saturation in the IMU, we find including the orientation estimates in the graph improves the robustness of the estimator. The estimator is updated at 1 kHz and optimized at 30Hz using the latest motion capture measurement and pre-integrated IMU measurements~\cite{imu_preintegration}. This system provides high quality, low latency state information suitable for control of a highly dynamic mechanism. In particular, we find that accurate linear velocity estimates enhance the performance of dynamic balancing controllers, such as the wheelie and hopping controllers.

\subsection*{Reinforcement Learning}
In this section, we detail the reinforcement learning (RL) framework, pipeline, and methods used for this work. We provide a high-level description of the components employed and how they can be reproduced.  Specifics of the RL algorithm, including reward design and parameter tuning, are detailed in~\cite{Kim2026}.

All of our controllers are \gls{rl}-based.
We aim to learn policies for \gls{umv} that generalize zero‐shot from simulation to real‐world.
The behaviors shown in the Results section are derived from four main policies:
\textit{drive}, 
\textit{wheelie-hop}, 
\textit{flip}, 
and \textit{jump}.
These policies are trained and deployed separately, 
but the \gls{rl} pipeline is the same with minor modifications depending on the goal of each policy.
While the driving, flipping, and hopping behaviors are proprioceptive, the jumping behavior is perceptive (e.g., moves automatically in reaction to the table).

\subsubsection*{Problem Formulation}
We frame the \gls{rl} problem as a~\gls{cmdp}, which is formally described by the tuple $(\mathcal{S}, \mathcal{A}, \mathcal{P}, r, \mathcal{C}, \gamma)$.
In this formulation:
$\mathcal{S}$ represents the state space, 
$\mathcal{A}$ is the action space, 
$\mathcal{P}(s_{t+1}|s_t, a_t)$ defines the probability of transitioning to state $s_{t+1}$ from state $s_t$ after taking action $a_t$, 
$r(s_t, a_t)$ is the reward function, 
$\mathcal{C} = \{c_1, ..., c_k\}$ denotes a set of $k$ distinct cost functions for the constraints, 
$\gamma \in [0, 1)$ is the discount factor.
The primary objective is to learn an optimal policy, $\pi^*$, that maximizes the expected cumulative discounted reward while strictly adhering to a set of operational constraints, $c_i(s_t, a_t) \leq 0$.
These constraints codify the physical limitations of the robotic hardware.
The optimization problem is stated as:
\begin{equation}
\pi^* = \arg\max_{\pi} \mathbb{E}_{\xi \sim \pi} \left[ \sum_{t=0}^{T} \gamma^t r(s_t, a_t) \right]
\end{equation}
\begin{equation}
\text{subject to } \quad c_i(s_t, a_t) \leq 0, \quad \forall i \in \{1,...,k\}, \forall t
\end{equation}

Employing a Constraints as Terminations approach~\cite{chane2024cat}, we handle these constraints by immediately terminating an episode upon any violation. This effectively converts the~\gls{cmdp} into a standard MDP solvable with \gls{rl}. With a reward structure that provides positive mean rewards at each step, any policy that violates a constraint will naturally result in a lower expected return, thus incentivizing the agent to learn safe behaviors. We employ Proximal Policy Optimization (PPO)~\cite{Rudin2022, schulman2017proximal} to compute policies.

\subsubsection*{Simulation and Control Loop}
Training is conducted within the \ig simulation environment~\cite{mittal2023orbit}. The RL-trained control policies operate at \unit[50]{Hz}, producing target setpoints for low-level PD controllers at each actuated DoF. These low-level controllers run at a higher frequency of \unit[200]{Hz} in simulation and \unit[8]{kHz} on the physical robot.

\subsubsection*{Joint and Action Spaces} The representation of the robot used during policy training in this work consists of five actuated DoFs or joints ($q$ = [$q_h$, $q_l$, $q_r$, $q_s$, $q_w$] as shown in~\fref{fig:kinematics}). Some policies, such as \textit{jump} and \textit{flip}, are best conceptualized as being purely sagittal: the~$q_l$ and~$q_r$ joints operate together as a single joint, locking the~$\phi$ angle at zero. We refer to this as the \textit{locked} representation. Therefore, while the joints and actions of the more general representation are denoted by~$q \in \mathbb{R}^5$ and~$a \in \mathbb{R}^5$, respectively, the joints and actions of the \textit{locked} representation are denoted by~$q \in \mathbb{R}^4$ and~$a \in \mathbb{R}^4$, respectively. Each action is scaled and then used as a PD target for a specific robot joint. The PD control law is implemented as \begin{equation} \tau = k_p(q_{\text{des}}- q) + k_d(\dot{q}_{\text{des}} - \dot{q}) \end{equation} where $q_{\text{des}}$ and $\dot{q}_{\text{des}}$ are the desired joint positions and velocities derived from the action vector~$a$, and $\tau$ is the resulting torque setpoint commanded by the motor driver. When commanding the rear wheel ($q_w$), $k_p$ is set to zero and the scaled action corresponds to a velocity setpoint. When commanding the other joints, the velocity setpoint is set to zero and the scaled action corresponds to a position setpoint.

\subsubsection*{Observations}
A common vector of proprioceptive observations $o_c \in \mathcal{S}$, provided to all policies is composed of: \begin{equation} o_c = [g, \omega, q, \dot{q}, \phi, \dot{\phi}, \mu, \dot{\mu}, a_{t-1}] \end{equation} where: $g \in \mathbb{R}^3$ is the gravity vector projected into the robot's \sit{Bike} frame, $\omega \in \mathbb{R}^3$ is the angular velocity of the \sit{Bike} frame, $q$ is the vector of measured joint positions, $\dot{q}$ is the vector of measured joint velocities, and $a_{t-1}$ is the action taken at the previous timestep. We also directly measure the angles $\phi$ and $\mu$ using encoders. Note that the position of the rear wheel is excluded from the observation space because it is an unbounded, continuously rotating joint. The \textit{drive} policy has an extra observation which is the commanded forward velocity and heading of the \sit{Bike}~$v_c~\in~\mathbb{R}^{2}$.

The \textit{jump} policies have two extra observations: a commanded goal location on the terrain in the world frame as seen from above~$\in~\mathbb{R}^{2}$, and a vector of robot-relative heights observing a local grid on terrain elevations in the direction of travel~$\in~\mathbb{R}^{112}$. The \textit{flip} policy has an extra observation, the phase variable~$\Theta \in \mathbb{R}^1$, that increases linearly from 0 to 1 over the duration of a reference trajectory. 

The \textit{wheelie-hop} policy has two extra observation commands~$\in~\mathbb{R}^{3}$. The first is a commanded forward velocity and heading of the \sit{Bike}~$v_c~\in~\mathbb{R}^{2}$. The second observation is the commanded gait mode~$g \in \mathbb{R}^1$, that determines whether the robot is driving, wheelieing, or hopping.

\subsubsection*{Reward Function Design}
The total reward is a weighted sum of reward terms designed to encourage accomplishing certain tasks and penalizing undesirable behaviors. The reward terms used, which depend on the behavior, are broadly categorized into three groups: \textit{task rewards}, \textit{style rewards}, and \textit{regularization rewards}. Reward terms are formulated as either the \textit{exponential kernel function}, \textit{squared exponential kernel function}, or \textit{squared function}, represented by Equations~\ref{eq:exponential_kernel}, \ref{eq:squared_exponential_kernel}, and \ref{eq:squared_function} below, respectively, where $x$ represents the input value, $x^c$ is the reference, and $\sigma$ is the standard deviation.
\begin{equation}
	r = \exp\left( - \frac{\left\|\mathbf{x} - \mathbf{x}^c \right\|}{\sigma^2} \right)
	\label{eq:exponential_kernel}
\end{equation}

\begin{equation}
	r = \exp\left( - \frac{\left\|\mathbf{x} - \mathbf{x}^c \right\|^2}{\sigma^2} \right)
	\label{eq:squared_exponential_kernel}
\end{equation}

\begin{equation}
	r = \left\|\mathbf{x} - \mathbf{x}^c \right\|^2
	\label{eq:squared_function}
\end{equation}

\textit{Task rewards} are the rewards with the highest weights and are meant to encourage the robot to accomplish a certain task. An example of a task reward is one that encourages the robot to track a certain commanded forward velocity.
\textit{Style rewards} encourage certain aesthetic qualities in the robot's behavior. An example of a style reward is one that encourages the robot to maintain a certain body configuration.
\textit{Regularization rewards} are used to avoid overfitting to training and to reduce unnecessary behaviors. An example of a regularization reward is one that penalizes torques.
The goal of the \textit{drive} policy is to track desired forward linear and angular yaw velocities (coming from a joystick) as the robot moves across flat ground. For these two tasks rewards it uses the exponential kernel function. During training of the \textit{jump} policy, the robot is given a starting location inside of a pit and a target location outside of the pit. As such, jumping is implicit rather than explicitly specified.  It has two main task rewards. The first reward is to reach the aforementioned goal location on the terrain. This tracking is measured using the squared Euclidean distance between the current position of the \sit{Bike} frame and the desired location. The second reward encourages the robot to orient its velocity vector toward the desired location. Both rewards are implemented using the squared exponential kernel function. The \textit{flip} policy attempts to imitate a reference generated by a model-based controller, that is kinematically and dynamically infeasible, while still respecting the robot's physical limits. Thus, it uses an imitation-based reward, wherein the robot is encouraged to track a reference trajectory for joint positions and the position of the \sit{Bike} frame. The \textit{flip} task reward uses the exponential kernel function.

The goal of the \textit{wheelie-hop} policy is to locomote in one of three modes: driving, wheelieing, and hopping. The policy includes a joystick command which designates the desired mode, though it has one main common task reward that encourages tracking the linear forward velocity and angular yaw velocity, regardless of the gait mode. Additionally, the policy includes extra task rewards that are conditioned on the gait mode~$g$ (drive, wheelie, or hop), which we refer to as \textit{gait-conditioned rewards}. Further detail on gait-conditioned rewards can be found in Siekmann et al~\cite{Siekmann2021}. Driving mode is similar to \textit{drive} above. In wheelieing mode, the policy encourages the front and rear wheels to remain off and on the ground, respectively. In hopping mode, the policy encourages the rear wheel to hop on the ground, depending on a pre-defined gait phase pattern (an alternating flight and stance phase). It also encourages the front wheel to avoid ground contact. When a wheel is in its stance phase, the policy is penalized for non-zero vertical wheel contact velocity. When a wheel is in its flight phase, the policy is penalized for non-zero vertical contact force at that wheel. The style and regulation rewards vary across different behaviors, and may include some or all of the following components.
A \textit{base orientation penalty} discourages the robot from deviating excessively from a desired orientation, such as leaning too far forward, backward, or sideways.
A \textit{high velocity penalty} is applied to prevent the robot from moving too quickly, while a \textit{yaw rate penalty} discourages excessive rotation around the vertical axis.
\textit{Joint position penalties} are used to avoid reaching the upper and lower limits of the robot's joint ranges, complemented by a separate joint position reward that encourages maintaining a target body configuration.
To ensure smooth and stable motion, \textit{action smoothness penalties} penalize large instantaneous changes in the processed network action outputs.
Additionally, \textit{joint torque and joint velocity penalties} discourage high torques or velocities that could stress the motors or destabilize the robot, and a \textit{contact force penalty} discourages excessive forces at the wheel-ground or body-ground contacts.
A \textit{jitter penalty} further penalizes high-frequency oscillations in the motion, promoting smoother trajectories.
A \textit{keep-alive reward} incentivizes the robot to avoid early terminations. Not all of these rewards are applied to every behavior; some are specific to certain joints while others act globally across all joints, and the corresponding weights of each reward may vary. Since the detailed tuning of these rewards is beyond the scope of this article, we present this summary for brevity.

\subsubsection*{Safety through Hard Constraints and Policy Limits} To guarantee safe operation on the physical robot, we enforce critical physical limits as hard constraints. We use episode termination as the mechanism for enforcing these constraints, a strategy also used by Chane-Sane et al~\cite{chane2024cat}. Instead of using soft termination constraints, which might allow a policy to learn recovery behaviors, we use hard termination constraints because any constraint violation in this context (e.g., excessive landing velocity or motor current) represents an immediate, unrecoverable failure that could lead to hardware damage. An episode is terminated if any of several safety or performance limits are violated. Specifically, termination occurs if the vertical landing speed of a wheel exceeds a safe threshold (\textit{touchdown velocity}), if the total mechanical power $\left(\sum \tau \dot{q}\right)$, which serves as a proxy for electrical current draw, surpasses a predefined limit (\textit{mechanical power}), or if a joint reaches its physical position limit, helping to prevent self-collisions (\textit{joint position limits}). Episodes are also ended if a motor's torque command exceeds its maximum rating (\textit{motor torque limits}), if the velocity of a wheel surpasses a safe operational speed (\textit{wheel velocity limit}), if the \sit{Bike} position or orientation deviates excessively from the reference motion (\textit{trajectory deviation}), or if any part of the robot's main body collides with the ground (\textit{ground collision}). To prevent the learning process from stalling due to these strict constraints, we introduce them gradually using a curriculum. Training starts with wider, more permissive constraint boundaries that are progressively tightened as the policy improves. This allows the agent to first learn the core task before refining its behavior to operate within the narrow safety margins of the hardware. Similar to the rewards, these termination terms vary across different behaviors, and may include some or all of the aforementioned components.

\subsubsection*{Domain Randomization}
To bridge the gap between simulation and reality, we use extensive domain randomization during all training phases. We randomize physical parameters such as mass, friction coefficients, center of mass locations, and motor strength. We also randomize control parameters such as PD gains and actuation delay. In addition, we inject noise into the observations (joint positions/velocities, \sit{Bike} angular velocity, gravity vector, and height map) and apply random external forces to the robot's body. Similar to the rewards and terminations, the domain randomization terms vary across different behaviors, and may include some or all of the aforementioned components.

\subsubsection*{Network Architecture}
Both the actor (policy) and critic (value function) are implemented as Multi-Layer Perceptrons (MLPs) with ELU activation functions.
The actor network uses three hidden layers with sizes [512, 256, 128], while the critic network's hidden layers are sized [512, 512, 256].
During training, the critic has access to privileged information—the \sit{Bike} link's linear velocity and global position—which is not available to the actor.
Depending on the policy, training takes between 15,000 and 42,000 iterations.

\clearpage
\section*{Acknowledgments}
The authors would like to thank Abel Gawel, Aleksandrs Ecins, Ananya Agarwal, Elliott Rouse, Jessica Hodgins, Katie Hughes, Luca Buonocore, Marco Zenti, Petr Listov, and Victor Reijgwart for their technical contributions, dedicated support, and insightful discussions throughout the development and testing of the \gls{umv} platform.  We also acknowledge the excellent interns who contributed to the UMV project: Abby Speckhals, Alexandre Amice, Antoine Leeman, Ardalan Tajbakhsh, Ashley Su, Catrina Sarno, Chris Evagora, Jeonghwan Kim, Ryan Grummer, and Seungeun Rho.\\
\textbf{Funding}: This research is supported by the Robotics and AI (RAI) Institute, which, in turn, receives funding from Hyundai Motor Group. The opinions expressed herein are those of the authors and independent of the views of the funding organizations.\\ 
%%%%%%%%%%%%%%%%%%%%%%%%%%%%%%%%%%%%%%%%%%%%%%%%%%%
%%%%%%%%%%%%%%% AUTHOR CONTRIBUTIONS %%%%%%%%%%%%%%
%%%%%%%%%%%%%%%%%%%%%%%%%%%%%%%%%%%%%%%%%%%%%%%%%%%
% BB lead the mechanical design and organized the writing.
% DJG did early work, but left the company eights months prior to initial submission. 
% SpnS [https://orcid.org/0000-0001-8112-4547] edited the manuscript and derived the 8 m/s result. 
%
\textbf{Author contributions}: The \textit{Contributors} developed and implemented the design optimization pipeline, designed and assembled the \gls{umv} robot, developed the embedded and training-tool software, trained most of the policies, and oversaw the hardware experiments on robots. They also contributed to the writing of the manuscript. They assisted with engineering tasks to support the project, including deployment code on hardware, data collection and curation, as well as supporting hardware experiments.  The \textit{Project Leads} set the overarching vision by defining the core scientific goals and managed the project. \\ 
\textbf{Competing interests}: The authors attest that there are no competing interests to declare. \\
\textbf{Data and materials availability:} Data to support the paper's conclusions are present herein or the Supplementary Materials.

\bibliography{UMV_System_Design} 

\clearpage\newpage

%%%%%%%%%%%%%%%%%%%%%%%%%%%%%%%%%%%%%%%%%%%%%%%%%%%%%%%%%%%%%%%%%%%%%%%%%%%%%%%%%%%%
%%%%%%%%%%%%%%%% START OF SUPPLEMENT %%%%%%%%%%%%%%%%%%%%%%%%%%%%%%%%%%%%%%%%%%%%%%%
%%%%%%%%%%%%%%%%%%%%%%%%%%%%%%%%%%%%%%%%%%%%%%%%%%%%%%%%%%%%%%%%%%%%%%%%%%%%%%%%%%%%

% Figures, tables, equations and pages in the supplement are numbered S1, S2 etc.
\renewcommand{\thefigure}{S\arabic{figure}}
\renewcommand{\themovie}{S\arabic{movie}}
\renewcommand{\thetable}{S\arabic{table}}
\renewcommand{\theequation}{S\arabic{equation}}
\renewcommand{\thepage}{S\arabic{page}}
\setcounter{figure}{0}
\setcounter{table}{0}
\setcounter{equation}{0}
\setcounter{movie}{0}
\setcounter{page}{1} 
% References continue the numbering from the main text.

\def\scititle{System Design of the Ultra Mobility Vehicle: \\A Driving, Balancing, and Jumping Bicycle Robot}

\section*{Supplementary Materials for\\ \scititle}

% Authors  
\author{Benjamin Bokser}, 
\author{Daniel J. Gonzalez}, 
\author{Aaron Preston}, 
\author{Alex Bahner}, 
\author{Annika Wollschläger}, 
\author{Arianna Ilvonen}, 
\author{Asa Eckert-Erdheim}, 
\author{Ashwin Khadke}, 
\author{Bilal Hammoud}, 
\author{Dean Molinaro}, 
\author{Fabian Jenelten}, 
\author{Henry Mayne}, 
\author{Howie Choset}, 
\author{Igor Bogoslavskyi}, 
\author{Itic Tinman}, 
\author{James Tigue}, 
\author{Jan Preisig}, 
\author{Kaiyu Zheng}, 
\author{Kenny Sharma}, 
\author{Kim Ang}, 
\author{Laura Lee}, 
\author{Liana Margolese}, 
\author{Nicole Lin}, 
\author{Oscar Frias}, 
\author{Paul Drews}, 
\author{Ravi Boggavarapu}, 
\author{Rick Burnham}, 
\author{Samuel Zapolsky}, 
\author{Sangbae Kim}, 
\author{Scott Biddlestone}, 
\author{Sean Mayorga}, 
\author{Shamel Fahmi}, 
\author{Surya P. N. Singh}, 
\author{Tyler McCollum}, 
\author{Velin Dimitrov}, 
\author{William Moyne}, 
\author{Yu-Ming Chen}, 
\author{David Perry}, 
\author{Farbod Farshidian}, 
\author{Marco Hutter}, 
\author{Al Rizzi}, and 
\author{Gabe Nelson}

\subsubsection*{This PDF file includes:}

\begin{itemize}
    \setlength{\itemsep}{0pt}
    \setlength{\parskip}{0pt}
    \item Fig S1.
A reduced 2D model used for vertical jump height optimization
    \item Fig S2.
An early study of the effect of gross robot scale on clearance height and contact ratio
    \item Fig S3.
Communications and power architecture
    \item Fig S4.
Sensitivity of balance controllability to design parameter changes for an extra DoF
    \item Table S1.
Nomenclature
    \item Table S2. Actuator specifications for \gls{umv}
\end{itemize}

\subsubsection*{Other Supplementary Materials for this manuscript:}
\begin{itemize}
    \setlength{\itemsep}{0pt}
    \setlength{\parskip}{0pt}
    \item Movie S1.
Continuous autonomous table jumping
    \item Movie S2.
Initial self-contained field demonstrations
\end{itemize}

\begin{figure}[h!]
    \centering
    \includegraphics[width=1.0\linewidth]{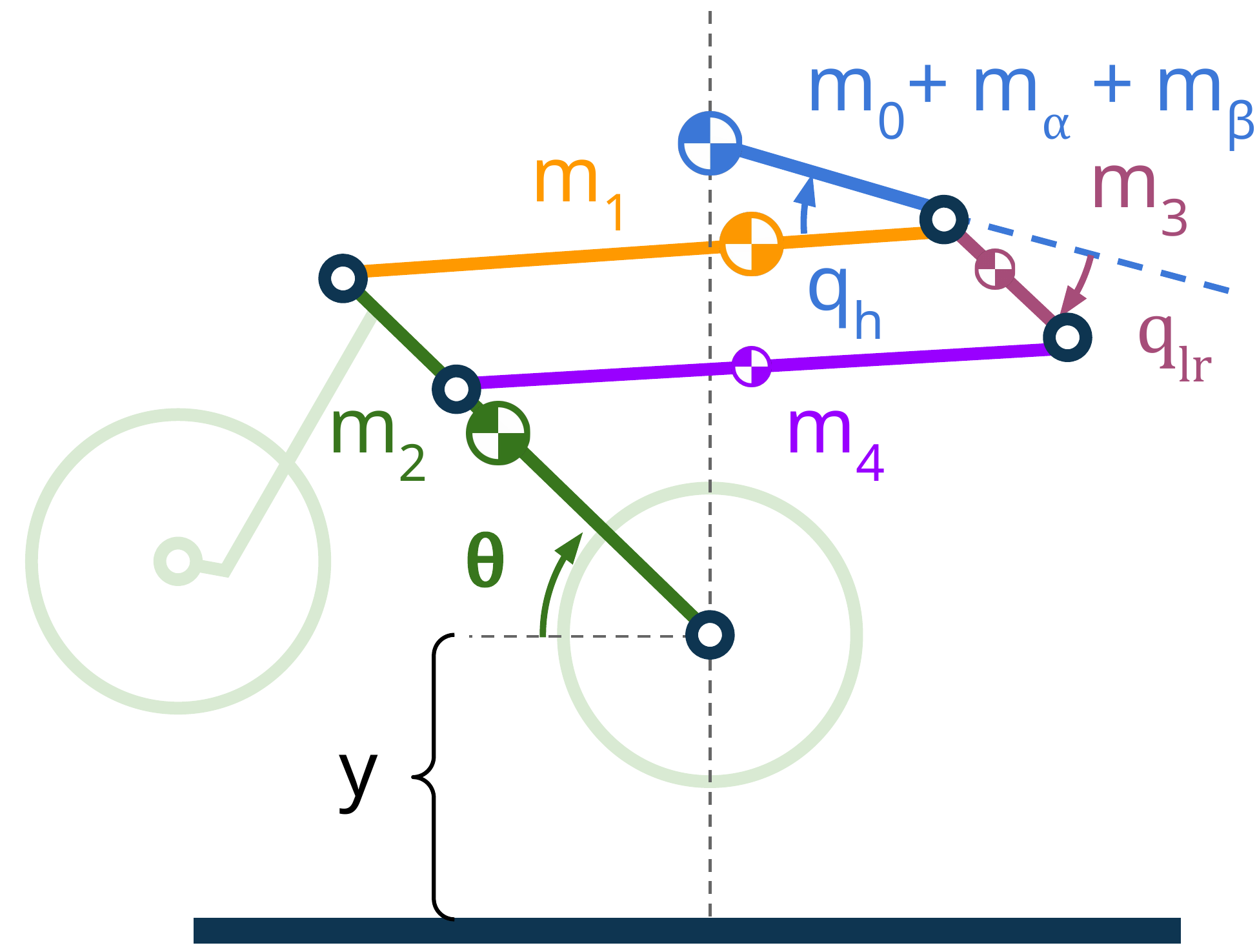}
    \caption{\textbf{A reduced 2D model used for vertical jump height optimization.}}
    \label{fig:diagram_2d}
\end{figure}

\begin{figure}[h!]
    \centering
    \includegraphics[width=1.0\linewidth]{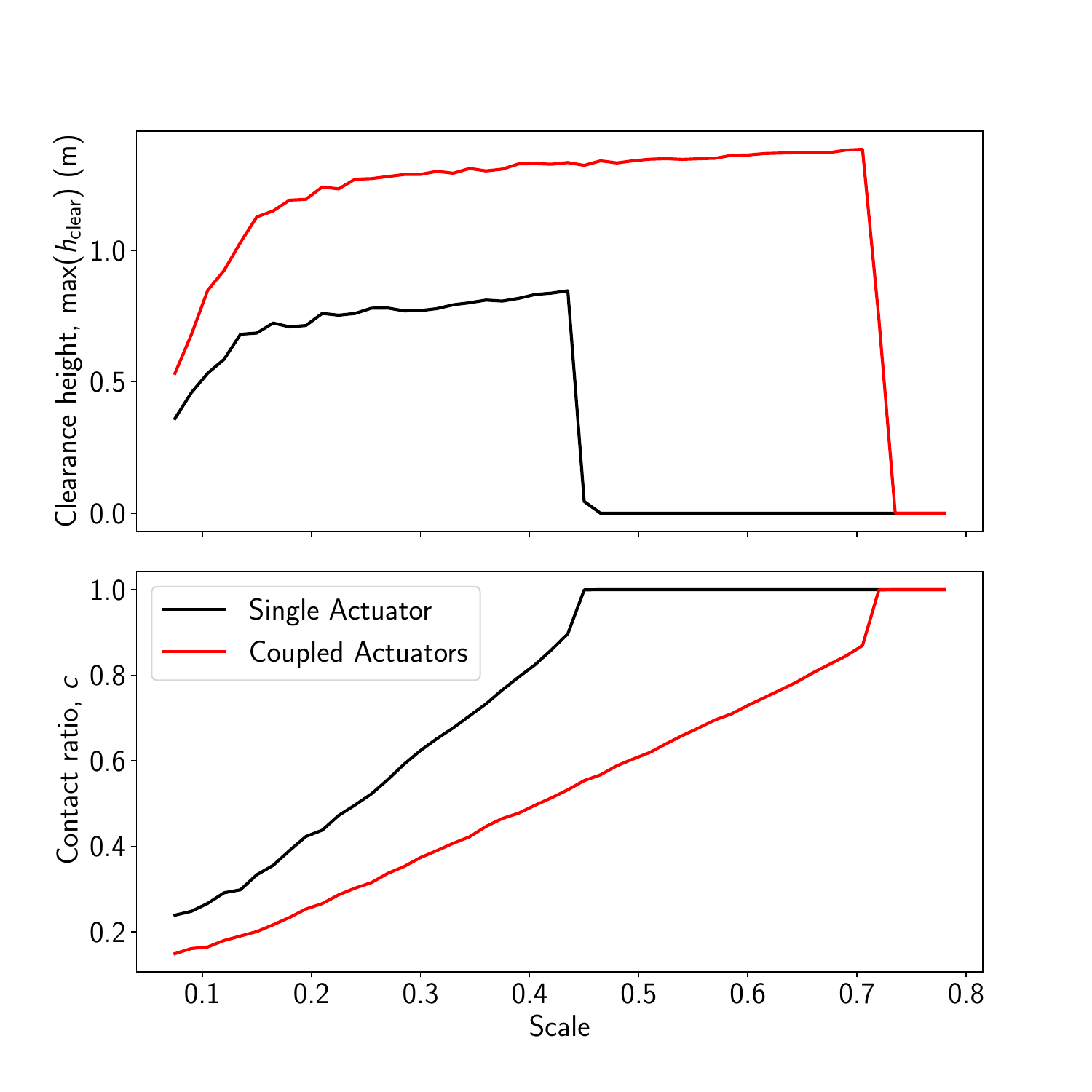}
    \caption{\textbf{An early study of the effect of gross robot scale on clearance height and contact ratio.} A ``scale'' of 1.0 corresponds to a robot with a \sit{Neck} length of 1 meter.
    With increasing scale, jump height increases up until the actuators are no longer able to provide the torque required to build upward momentum.
    We compared the use of a single AK10-9 actuator to drive joint $q_h$ versus the use of a set of coupled AK10-9 actuators for the same purpose.
    This study used the actuator's original gear ratio of 9:1, and the relationship between link length, mass, and moment of inertia was determined by approximating links as cylinders.}
    \label{fig:coupled}
\end{figure}

\begin{figure}[h!] % Do NOT use \begin{figure*}
	\centering
	\includegraphics[width=1.\columnwidth]{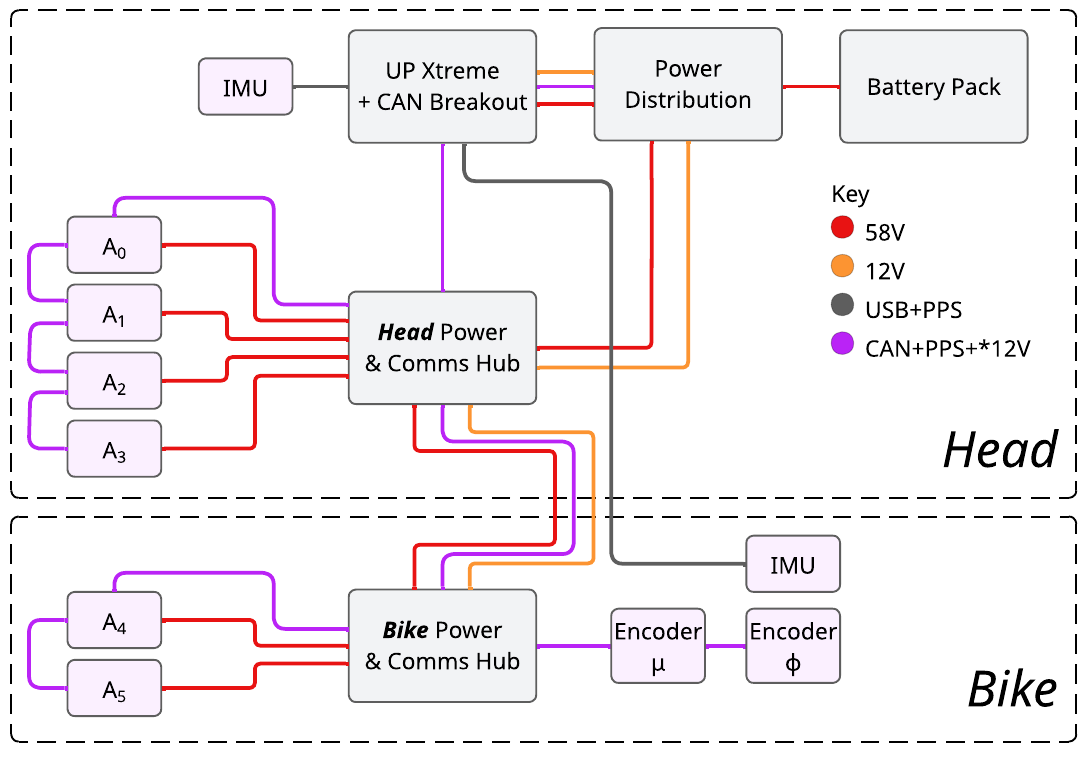}
	\caption{\textbf{Communications and power architecture.} All actuators ($A_0$ through $A_5$) have encoders built-in.
    More information on the mapping of actuators to joints can be seen in Table \ref{tab:actuators}.
    Because each \gls{rl} policy requires observations of one of the two bodies, we mount one IMU on the \sit{Head} and another on the \sit{Bike} link.
    With only one IMU, the angular and linear velocities of the other body would become a function of the (noisy) derivative of the joint position measurements, decreasing the overall accuracy of the observations.}
	\label{fig:electronics}
\end{figure}

\begin{figure*}[htb]
    \centering
    \includegraphics[width=1.0\textwidth]{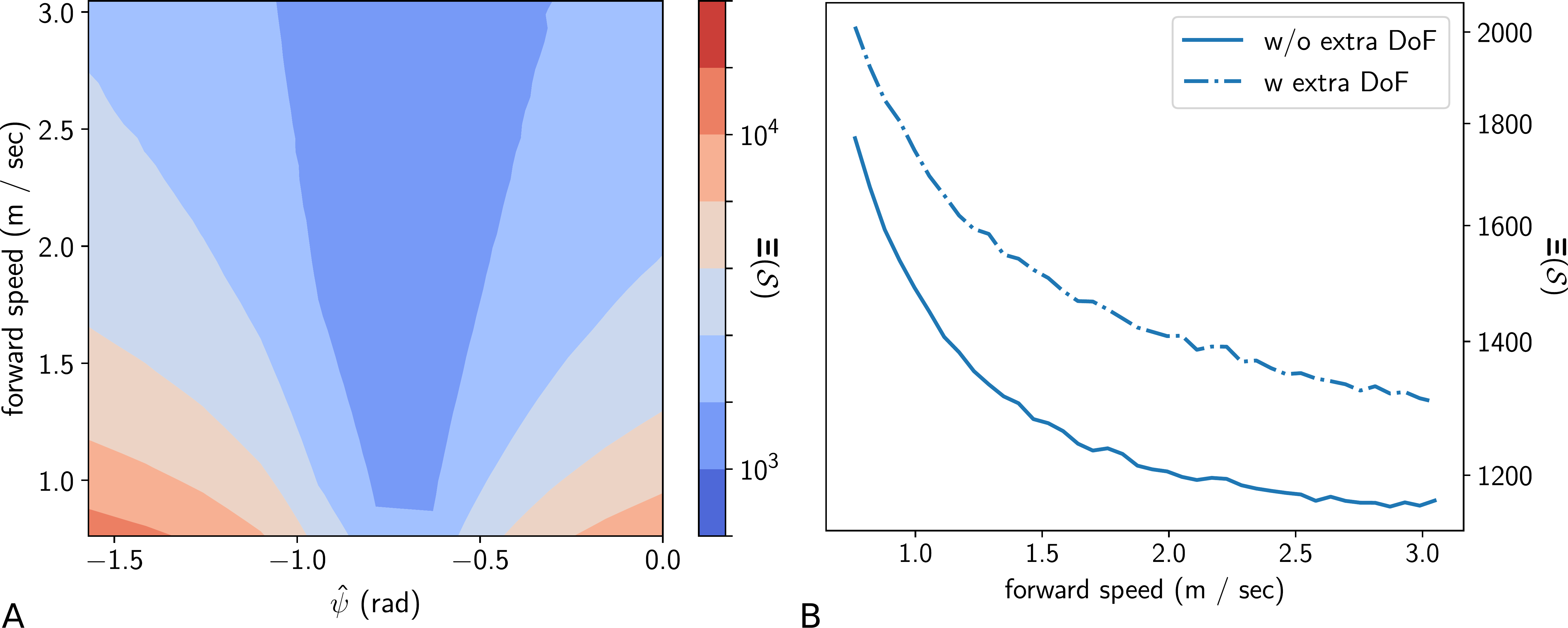}
    \caption{\textbf{Sensitivity of balance controllability to design parameter changes for an extra DoF.} (\textbf{A}) Variation in minimum required control effort (\eqref{eq:aggregate_controllability}) during wheelie balance with variation of the axis orientation $\hat{\psi}$ of joint $\phi$ as depicted in \fref{fig:kinematics}.
    The optimal value of $\hat{\psi}$ is such that the axis of joint $\phi$ roughly intersects the rear axle.
    (\textbf{B}) Comparison of the minimum required control effort of the current morphology (without extra DoF) to that of a 6-DoF morphology (with extra DoF) shows that it would require additional control effort in balancing and is not necessary.}
    \label{fig:yoll_controllability_opt}
\end{figure*}

%%%%%%%%%%%%%%%% SUPPLEMENTARY TABLES %%%%%%%%%%%%%%%
\begin{table}[h!] % Do NOT use \begin{table*}
	\centering
	\caption{\textbf{Nomenclature}.}
	\label{tab:nomenclature} % give each table a logical label name
    \renewcommand{\arraystretch}{0.9} % Adjust this value to control row spacing
	\begin{tabular}{cl}
        $A_i$ & Actuators \\
        $\omega$ & Angular velocity of the bike frame \\
        $q_h$ & Joint between the \sit{Head} and \sit{Neck} links \\
        $q_l$ & Right arm joint \\
        $q_r$ & Left arm joint \\
        $q_s$ & Steering joint \\
        $q_w$ & Rear wheel joint \\
        $\boldsymbol{q}$ & Driven joint positions vector \\
        $\boldsymbol{\dot{q}}$ & Driven joint velocities vector \\
        $\boldsymbol{q_\text{des}}$ & Desired driven joint positions vector\\
        $\mu$ & Pitch joint of the \sit{Neck} link \\
        $\phi$ & Out-of-plane joint \\
        $\theta$ & Front wheel joint, passive \\
        $\hat{\psi}$ & Fixed angle of the axis of $\phi$ \\
        $\hat{\zeta}$ & Fixed angle of the axis of alternative $\phi$ \\
        $p_\text{CoM}$ & Position vector of the whole-body CoM \\
        $h_\text{CoM}$ & Height of the whole-body CoM \\
         $p_\text{clearance}$ & Position vector of rear wheel's lowest point \\
        $h_\text{clearance}$ & Height of rear wheel's lowest point \\
        $\max(h_\text{CoM})$ & CoM height at apogee \\
        $\max(h_\text{clearance})$ & Clearance height at apogee \\
        $c$ & Contact ratio \\
        $\mathcal{S}$ & Bounded state space \\
        $u$ & Control inputs \\
        $x$ & Vector of state variables \\
        $x_s$ & Starting state \\
        $x_f$ & Final state \\
        $\boldsymbol{\Xi}(\mathcal{S})$ & Minimum control effort metric \\
        $\boldsymbol{E}$ & Effort \\
        $r$ & Reward \\
        $a$ & Action vector \\
    \end{tabular}
\end{table}
% \newpage
\begin{table}[h!] % Do NOT use \begin{table*}
	\centering
	% \label{tab:nomenclature} % give each table a logical label name
    
 \renewcommand{\arraystretch}{0.9} % Adjust this value to control row spacing
	\begin{tabular}{cl}
        $k_p$, $k_d$ & PD gains \\
        $\tau$ & Torque setpoint command \\
        $v_c$ & Base forward vel. and heading command \\
        $\Theta$ & Phase variable \\
        $g$ & Gravity vector \\
        $o_c$ & Observation vector \\
        $a$ & Actions vector \\
        $\mathcal{S}$ & State space \\
        $\mathcal{A}$ & Action space \\
        $\mathcal{P}$ & Probability \\
        $\mathcal{C}$ & Cost Functions \\
        $\gamma$ & Discount factor \\
	\end{tabular}
\end{table}

\begin{table}[h!] % Do NOT use \begin{table*}
	\centering
	\caption{\textbf{Actuator Specifications for \gls{umv}.}
		See~Fig.~\ref{fig:kinematics}B for placement on the robot.}
	\label{tab:actuators} % give each table a logical label name
    \setlength{\tabcolsep}{3pt} % narrows table to fit in text column, remove before submission
	\begin{tabular}{lccccc} % five columns, alignment for each
		\\
        \hline
        Actuator & P/N & Gear & Max Torque & Max Speed & Driven \\
		  &  & Ratio & (Nm) & (rad/s) & Joint(s)\\
		\hline
		$A_0$ & AK10-9 & 297:22 & 72 & 
 29.4 & $q_r$\\
        $A_1$ & AK10-9 & 297:22 & 72 & 29.4 & $q_l$\\
		$A_2$ + $A_3$ & AK10-9 & 450:22 & 218 & 19.4 & $q_h$\\
        $A_4$ & R80 & 13:2 & 26 & 74 & $q_w$\\
        $A_5$ & R60 & 38:14 & 6.25 & 160 & $q_s$\\
		\hline
	\end{tabular}
\end{table}

%%%%%%%%%%%%%%%% SUPPLEMENTARY MOVIES %%%%%%%%%%%%%%%
%%%%%%%%%%% CAPTIONS FOR OTHER SUPPLEMENTARY FILES %%%%%%%%%%

\clearpage % Clear all remaining figures and tables then start a new page
\setcounter{figure}{0}

\makeatletter
\renewcommand{\fnum@figure}{Movie \thefigure}
\makeatother

%%%%%%%%%%%%%%%%%%%%%%%%%%%%%%%%%%%%%%%%%%%%%%%%%%%
% MOVIE: TABLE JUMP MOVIE 
%%%%%%%%%%%%%%%%%%%%%%%%%%%%%%%%%%%%%%%%%%%%%%%%%%%
\begin{movie*}[h]
    \centering
    \includegraphics[width=1.0\linewidth]{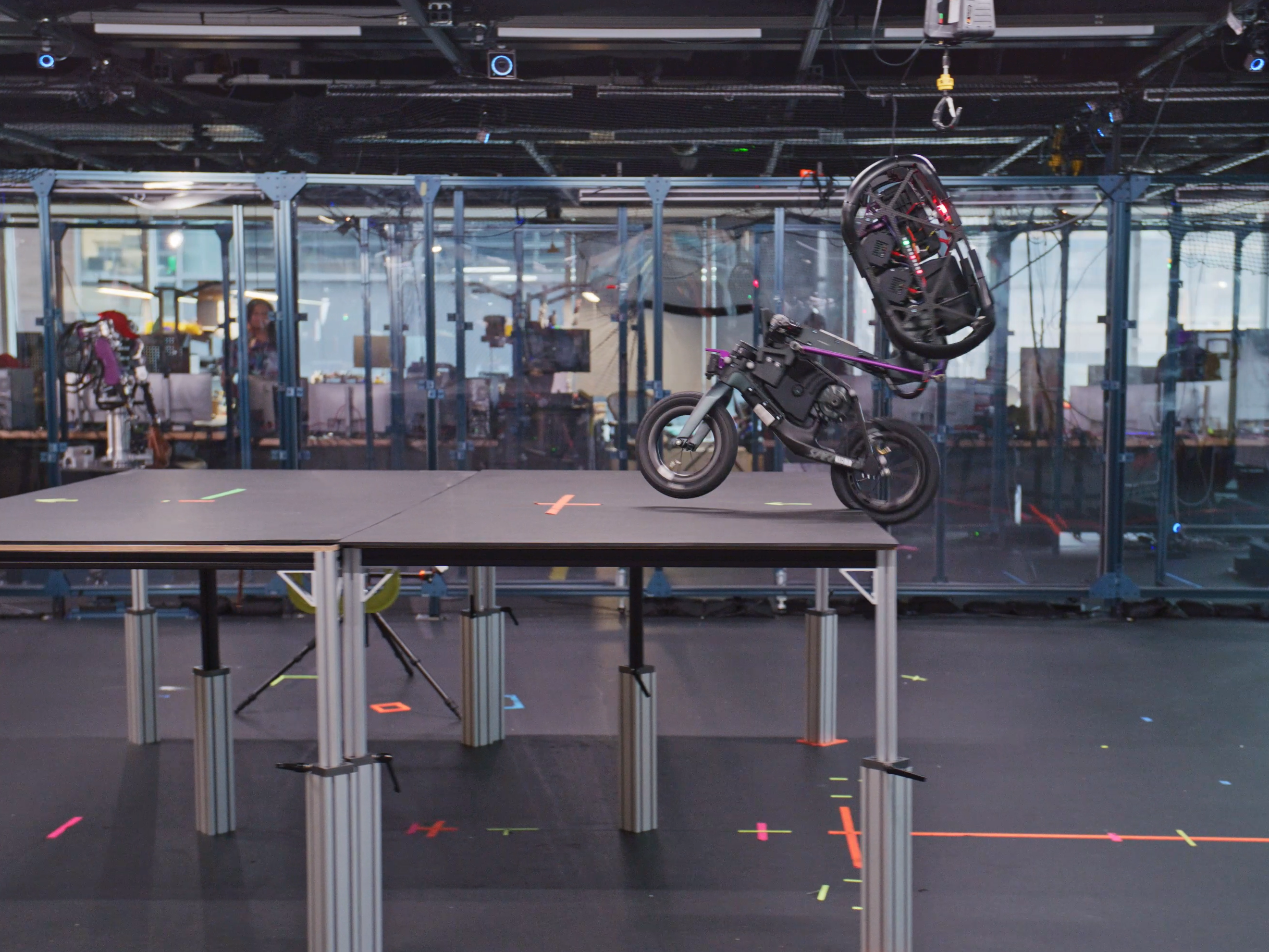}
    \caption{\textbf{Continuous autonomous table jumping.} 
    The robot executes a continuous routine of fifteen autonomous jumps onto and off of a 1-meter platform  without intervention.
    This experiment illustrates the policy's repeatability and the hardware's endurance under sustained loads.
    The trajectories of this motion are analyzed in Fig.~\ref{fig:table_jump}B.}
    \label{movie:tablejump}
\end{movie*}

%%%%%%%%%%%%%%%%%%%%%%%%%%%%%%%%%%%%%%%%%%%%%%%%%%%
% MOVIE: INITIAL FIELD DEMONSTATIONS
%%%%%%%%%%%%%%%%%%%%%%%%%%%%%%%%%%%%%%%%%%%%%%%%%%%
\begin{movie*}[h]
    \centering
    \includegraphics[width=1.0\linewidth]{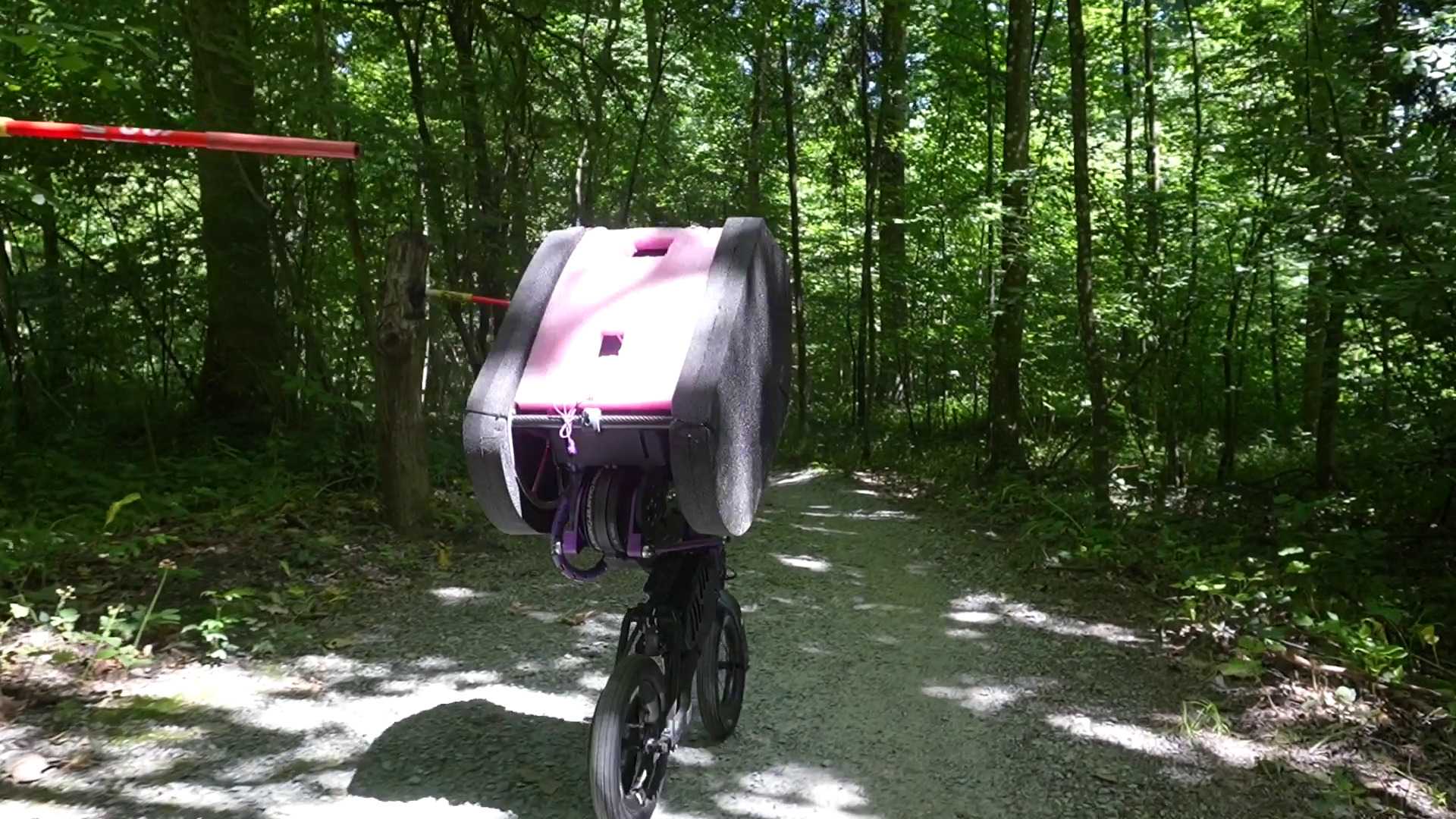}
    \caption{\textbf{Initial self-contained field demonstrations.} The robot is shown operating outdoors along a biking trail (Allmendtrail, Zürich).
    The \gls{umv} robot performs essential locomotion tasks --- including balancing, turning, and acceleration --- using fully onboard power, state estimation, and controls.}
    \label{movie:field}
\end{movie*}

\end{document}